\documentclass[opre,nonblindrev]{informs3re}

\DoubleSpacedXI 

\usepackage{endnotes}
\let\footnote=\endnote

\usepackage{graphicx}
\usepackage{subfigure, epsfig}
\usepackage{natbib}
\usepackage{clipboard}
\usepackage{epstopdf}

\bibpunct[, ]{(}{)}{,}{a}{}{,}%
\def\bibfont{\small}%
\TheoremsNumberedThrough     
\ECRepeatTheorems

\EquationsNumberedThrough    

\usepackage{amsmath}
\usepackage{amssymb}
\usepackage{mathtools}
\mathtoolsset{showonlyrefs}
\usepackage{bm}

\usepackage{algorithm}
\usepackage{algorithmic}

\usepackage{booktabs} 
\usepackage{comment}
\usepackage{float}
\usepackage{enumerate}
\newcommand{\R}{\mathbb{R}}

\newcommand{\E}{\mathbb{E}}

\newcommand{\B}{\mathcal{B}}
\newcommand{\I}{\mathcal{I}}
\newcommand{\Prob}{\mathbb{P}}
\usepackage{dsfont}
\newcommand{\1}{\mathds{1}}

\usepackage{hyperref}
\newcommand{\rev}[1]{\textcolor{black}{#1}}

\begin{document}
	
	
	\RUNAUTHOR{Xiong, Chen, Gao and Zhou}
	
	\RUNTITLE{POMDPs Learning}
	
	\TITLE{Sublinear Regret for Learning POMDPs}
	
	\ARTICLEAUTHORS{	
		\AUTHOR{Yi Xiong}
		\AFF{Department of Systems Engineering and Engineering Management, The Chinese University of Hong Kong, Hong Kong, China, \EMAIL{yxiong@se.cuhk.edu.hk}}
		\AUTHOR{Ningyuan Chen}
		\AFF{Rotman School of Management, University of Toronto, Toronto, Canada, \EMAIL{ningyuan.chen@utoronto.ca}}
		\AUTHOR{Xuefeng Gao\thanks{Corresponding author.}}
		\AFF{Department of Systems Engineering and Engineering Management, The Chinese University of Hong Kong, Hong Kong, China, \EMAIL{xfgao@se.cuhk.edu.hk}}
		\AUTHOR{Xiang Zhou}
		\AFF{Department of Systems Engineering and Engineering Management, The Chinese University of Hong Kong, Hong Kong, China, \EMAIL{zhouxng@se.cuhk.edu.hk}}
	} 
	
	\ABSTRACT{%
		We study the model-based undiscounted reinforcement learning for partially observable Markov decision processes (POMDPs). We propose a learning algorithm for this problem, building on spectral method-of-moments estimations for hidden Markov models, the belief error control in POMDPs and upper-confidence-bound methods for online learning. We establish a regret bound of $O(T^{2/3}\sqrt{\log T})$ for the proposed learning algorithm where $T$ is the learning horizon. This is, to the best of our knowledge, the first algorithm achieving sublinear regret for learning general POMDPs.
	}%
	
	
	
	\maketitle
	
	%
	
\section{Introduction}
The partially observable Markov decision process (POMDP) is a framework for dynamic decision-making when some evolving state of the system cannot be observed.
It extends the Markov decision process (MDP) and can be used to model a wide variety of real-world problems, ranging from healthcare to business.
The solution to POMDPs is usually through a reduction to MDPs, whose state is the belief (a probability distribution) of the unobserved state of the POMDP, see e.g. \cite{krishnamurthy2016partially} for an overview. 

We study the problem of decision making when the environment of the POMDP, such as the transition probability of the hidden state and the probability distribution governing the observation, is unknown to the agent.
Thus, the agent has to simultaneously learn the model parameters (we use ``environment'' and ``parameters'' interchangeably) and take optimal actions.
Such online learning framework has received considerable attention in the last decades \citep{sutton2018reinforcement}.
Despite the practical relevance of POMDPs, the learning of POMDPs is considered much more challenging than finite-state MDPs and few theoretical results are known.
This is not surprising:
even with a known environment, the corresponding belief MDP features a continuous state space.
When the environment is unknown, we face the additional difficulty of not being able to calculate the belief accurately, whose updating formula is based on the environment.
This is in contrast to the learning of standard MDPs, in which the state is always observed exactly.

To tackle this daunting task, we provide an algorithm that achieves sublinear regret, which is a popular measure for the performance of a learning algorithm relative to that of the oracle, i.e., the optimal policy in the known environment.
This is the first algorithm that achieves sublinear regret, to our knowledge, in the general POMDP setup we consider.
We summarize the \emph{three major contributions} of this paper below.

In terms of problem formulation, we benchmark our algorithm against an oracle and measure the performance by calculating the regret.
The oracle we consider is the strongest among the recent literature \citep{azizzadenesheli2016reinforcement,fiez2018multi}.
\Copy{rev:oracle}{\rev{In particular, the oracle is
		the optimal policy of the POMDP with a known environment in terms of the average reward over an infinite horizon.}}
Such an oracle has higher average reward than the oracles that use the best fixed action \citep{fiez2018multi} or the optimal memoryless policy (the action only depends on the current observation) \citep{azizzadenesheli2016reinforcement}.
Still our algorithm is able to attain sublinear regret in the length of the learning horizon.
This implies that as the learning horizon increases, the algorithm tends to approximate the strong oracle more accurately.

In terms of the algorithmic design, the learning algorithm we propose (see Algorithm \ref{alg:SEEU}) has two key ingredients.
First, it builds on the recent advance on the estimation of the parameters of hidden Markov models (HMMs) using spectral method-of-moments methods, which involve the spectral decomposition of certain low-order multivariate moments computed from the data \citep{anandkumar2012method, anandkumar2014tensor, azizzadenesheli2016reinforcement}.
It benefits from the theoretical finite-sample bound of spectral estimators,
while the finite-sample guarantees of other alternatives such as maximum likelihood estimators remain an open problem \citep{lehericy2019consistent}.\footnote{There are recent advances on the EM algorithm that are applied to likelihood-based methods for HMMs \citep{balakrishnan2017statistical,yang2017statistical} with finite-sample analysis. However, the conditions on the $Q$ function and the resulting basin of attraction are hard to translate to our setting explicitly.}
Second, it builds on the well-known
``upper confidence bound'' (UCB) method in reinforcement learning \citep{auer2006logarithmic, jaksch2010near}.
We divide the horizon into nested exploration and exploitation phases.
We use spectral estimators in the exploration phase to estimate the unknown parameters such as the transition matrix of the hidden state, which itself is a function of the action in the period.
We apply the UCB method to control the regret in the exploitation phase based on the estimated parameters in the exploration phase and the associated confidence regions.
Although the two components have been studied separately before, it is a unique challenge to combine them in our setting.
In particular, the belief of the hidden state is subject to the estimation error.
We re-calibrate the belief at the beginning of each exploitation phase based on the most recent estimate of the parameters.
This helps us achieve the sublinear regret.

In terms of regret analysis, we establish a regret bound of $O(T^{2/3} \sqrt{\log(T)})$ for our proposed learning algorithm where $T$ is the learning horizon. 
Our regret analysis draws inspirations from \cite{jaksch2010near, ortner2012online} for learning MDPs and undiscounted reinforcement learning problems, but the analysis differs significantly from theirs since there are two main technical challenges in our problem.

First, the belief in POMDPs, unlike the state in MDPs,
is not directly observed and needs to be estimated.
This is in stark contrast to learning MDPs \citep{jaksch2010near, ortner2012online} with observed states.
As a result, we need to bound the estimation error of the belief which itself depends on the estimation error of the model parameters.
In addition, we also need to bound the error in the belief transition kernel, which depends on the model parameters in a complex way via Bayesian updating. \rev{To control these errors,  we extend the approach in \cite{de2017consistent} for HMM to POMDP, and relate the error in belief transitions to the estimation error of POMDP parameters and the belief state error.}

Second, to establish the regret bound, we need an uniform bound for the span of the bias function (also referred as the relative value function) for the optimistic belief MDPs which have continuous state spaces.
Such a bound is often critical in the regret analysis of undiscounted reinforcement learning of continuous MDP, but it is often shown under restrictive assumptions such as
the H\" older continuity that do not hold for the belief state in our setting \citep{ortner2012online, lakshmanan2015improved}.
We develop a novel approach to bound the bias span for the undiscounted POMDP by bounding the Lipschitz modulus of the optimal value function for infinite-horizon discounted problems when the discount factor tends to one.
One key step is to bound the Lipschitz module of the belief transition kernels using the Kantorovich
metric. Exploiting the connection with the infinite-horizon undiscounted problem via the vanishing discount factor method then yields an explicit bound on the bias span for the optimisitic belief MDPs. 

\subsection{Related Literature}\label{sec:literature}

This paper extends the online learning framework popularized by multi-armed bandits to POMDPs.
There is a large stream of literature on the topic of bandits, see e.g. \citet{bubeck2012regret} for a survey.
In POMDPs, the rewards across periods are not independent any more.
A stream of literature studies nonstationary/switching MAB, including \citet{auer2002nonstochastic, garivier2011upper, besbes2014stochastic,keskin2017chasing,cheung2022hedging,auer2019adaptively}.
The reward can change over periods subject to a number of switches or certain changing budget (the total magnitude of reward changes over the horizon), and the oracle is the best action in each period.
It should be noted that the oracle considered is stronger than ours.
However, all the designed algorithms in this literature require finite switches or sublinear changing budget (in the order of $o(T)$).
This is understandable, as there is no hope to learn such a strong oracle if the actions can be completely different across periods.
In our setting, the number of changes (state transitions) is linear in $T$ and the algorithms are expected to fail to achieve sublinear regret even measured against our oracle, which is weaker than the oracle in this stream of literature.
There are a few exceptions, including \cite{zhu2020demands,chen2021learning,zhou2021regime}, which study models with linear changing budget but specific structures.
In \cite{chen2021learning}, the rewards are cyclic which can be leveraged to learn across cycles despite of the linear change.
In \cite{zhu2020demands}, the reward grows over time according to a function.
In \cite{zhou2021regime}, the reward is modulated by an unobserved Markov chain.
Another stream of literature investigates the so-called restless Markov bandit problem, in which the state of each action evolves according to independent Markov chains, whose states may not be observable.
See, for example, \cite{slivkins2008adapting,guha2010approximation, ortner2014regret}.
The POMDP model we consider has a more complex structure. Thus the algorithms proposed in the above studies cannot achieve sublinear regret.

Our work is related to the rich literature on learning MDPs.
\cite{jaksch2010near} propose the UCRL2 (Upper Confidence Reinforcement Learning) algorithm to learn finite-state MDPs and prove that the algorithm can achieve the optimal rate of regret measured against the optimal policy in terms of the undiscounted average reward.
Follow-up papers have investigated various extensions to \cite{jaksch2010near}, including posterior sampling \citep{agrawal2017optimistic}, minimax optimal regret \citep{azar2017minimax, zhang2019regret}, and the model-free setting \citep{jin2018q}.
\cite{cheung2019non} consider the case where the parameters of the MDP, such as the transition matrix, may change over time.
The algorithms are not applicable to our setting, because of the unobserved state in POMDPs.
However, since a POMDP can be transformed to a continuous-state MDP, our setting is related to the literature, especially those papers studying MDPs with a continuous state space.
\cite{ortner2012online,lakshmanan2015improved} extend the algorithm in \cite{jaksch2010near} to a continuous state space.
Still, our problem is not equivalent to the learning of continuous-state MDPs.
First, in this literature H\"older continuity is typically assumed for the rewards and transition probabilities with respect to the state, in order to aggregate the state and reduce it to the discrete case.
However, this assumption does not hold in general for the belief state of POMDPs, whose transition probabilities are not given but arise from the Bayesian updating.
Second, even if the continuity holds, the state of the belief MDP in our problem, which is the belief of the hidden state, cannot be observed.
It can only be inferred using the estimated parameters.
This distinguishes our problem from those studied in this literature.
The algorithm and analysis also deviate substantially as a result.
There are studies that focus on the applications such as inventory management \citep{zhang2018perishable,chen2019coordinating,zhang2020closing,chenshi2020optimal,nambiar2021dynamic} and handle specific issues such as demand censoring and lost sales.

Our work is related to studies on reinforcement learning for POMDPs, see e.g. \cite{ross2011bayesian,spaan2012partially} and references therein.  \cite{guo2016pac} propose a learning algorithm for a class of episodic POMDPs, where the performance metric is the sample complexity, i.e. the time required to find an approximately optimal policy. {Recently, \cite{jin2020sample} give a sample efficient
	learning algorithm for episodic finite undercomplete POMDPs, where the number of observations is larger than the number of hidden states.
	{Their focus is on the sample
		complexity, while the method may potentially be used in the regret analysis of our infinite-horizon
		average reward setting. }
	There is also a growing body of literature that apply deep reinforcement learning methods to POMDPs, see e.g. \cite{hausknecht2015deep,igl2018deep}.
	Our work differs from these papers in that we study the learning of ergodic POMDPs in an unknown environment and we focus on developing an learning algorithm with sublinear regret guarantees. {A concurrent study \citep{kwon2021rl} considers regret minimization for reinforcement learning in a special class of POMDPs called latent MDP. The hidden state is static in their work while it is dynamic in our setting.}
	
	Furthermore, our work is related to the literature on the spectral method to estimate HMMs and its application to POMDPs. For instance,
	\cite{anandkumar2012method,anandkumar2014tensor} use the spectral method to estimate the unknown parameters in HMMs, by constructing the so-called multi-views from the observations.
	The spectral method is not readily applicable to POMDPs, because of the dependence introduced from the actions.
	\cite{azizzadenesheli2016reinforcement} address the issue by restricting to memoryless policies, i.e., the action  only depends on the observation in the current period instead of the belief state.
	They extend the spectral estimator to the data generated from an arbitrary distribution other than the stationary distribution of the Markov chain, which is necessary in learning problems when the policy needs to be experimented.

	\label{page:lit-diff}
	\Copy{rev:literature}{\rev{
			There are two papers whose methodology is closely connected to this paper that warrant more discussion \citep{azizzadenesheli2016reinforcement,de2017consistent}.
			\cite{azizzadenesheli2016reinforcement} use spectral estimators and upper confidence methods to learn POMDPs and establish a regret bound of $O(\sqrt{T} \log(T))$.
			One main difference between our work and theirs is the choice of the oracle/benchmark.
			Specifically, their oracle is the optimal memoryless policy, i.e., a policy that only depends on the current reward observation instead of using all historical observations to form the belief of the underlying state.
			For general POMDPs, memoryless policies are suboptimal and the performance gap is linear in $T$ between their oracle and ours.
			Technically, it allows them to circumvent the introduction of the belief state entirely.
			In our setting, we need to design a new learning algorithm to achieve sublinear regret with our stronger oracle and analyze the regret.
			By considering the belief-based policies, several new difficulties arise in our setting.
			First, the spectral method can not be applied to samples generated from belief-based policies due to history dependency; Second, the belief states can not be observed and need to be calculated using the estimated parameters, which is not an issue in \cite{azizzadenesheli2016reinforcement} because the observation in the current period can be regarded as the state.
			We tackle these difficulties by using an exploration-exploitation interleaving approach in the algorithm design.
			In particular, we develop three recipes to analyze the regret.
			First, we develop a novel approach to upper bound the span of the bias function for the average-reward POMDP model.
			In general the bias span is a complicated function of the POMDP model parameters. Our paper appears to be the first to make the bound explicit in terms of the smallest element of the transition matrices, and this is one of our main methodological contributions. This result is significant as it simultaneously provides bounds on the bias span of the estimated POMDP and the optimistic POMDP when they are close to the true POMDP model in terms of model parameters.
			Our bound on the bias span is different from the diameter of the POMDP discussed in \cite{azizzadenesheli2016reinforcement}. The diameter in \cite{azizzadenesheli2016reinforcement} is only
			for observation-based policies, not for belief-state based policies we consider.
			Second, to control the regret, we bound the error in the belief state incurred by the errors in the estimation of POMDP parameters.
			We extend the approach in \cite{de2017consistent}, which study filtering and smoothing errors in the context of nonparametric hidden Markov models.
			Such HMM models do not involve actions or decision making as in the POMDP model we consider.
			Third, to control the regret, we also bound the error in the (estimated) belief transition law. This is similar to online learning of finite-state MDPs where one often bounds the error in the estimates of transition probabilities in model-based methods.
			However, our problem is more sophisticated because the belief state is not observed, and hence can not be directly estimated. Therefore, we control this error in the transition law of beliefs by relating it to the estimation error of POMDP parameters and the belief state error.
		}
	}

	
	The rest of the paper is organized as follows. In Section~\ref{sec:formulation} we discuss the problem formulation. Section~\ref{sec:alg} presents our learning algorithm. In Section 4,
	we state our main results on the regret bounds for the learning algorithm. \rev{In Section~\ref{sec:numerical}, we present numerical experiments.}
	Finally, we conclude in Section~\ref{sec:conclusion}. All the proofs of the results in the paper are deferred to the appendix.

	\section{Problem Formulation}\label{sec:formulation}
	We first introduce the notation for the POMDP.
	A POMDP model with horizon $T$ consists of the tuple
	\begin{align}\label{eq:pomdp-tuple}
		\{\mathcal{M} ,\mathcal{I} , \mathcal{O}, \mathcal{P}, \Omega, R \},
	\end{align}
	where
	\begin{itemize}
		\item $\mathcal{M}\coloneqq\{1,2,\dots,M\}$ denotes the state space of the hidden state.
		We use $M_t \in \mathcal{M}$ to denote the state at time $t=1, 2, \ldots, T$.
		
		\item $\mathcal{I}\coloneqq\left\{1, 2, \dots,I\right\}$ denotes the action space with $I_t \in \mathcal{I}$ representing the action chosen by the agent at time $t$.
		
		\item $\mathcal{O} \coloneqq\{o_1, o_2, \dots, o_O\} $ is a finite set of possible observations and $O_t$ denotes the observation at time $t$.
		
		\item  $\mathcal{P}\coloneqq\{P^{(1)},\dots,P^{(I)}\}$ describes a family of transition probability matrices, where $P^{(i)} \in \R^{M\times M} $ is the transition probability matrix for states in $\mathcal{M}$ after the agent takes action $i \in \mathcal{I}$. That is, $P^{(i)}(m, m') = \mathbb{P}(M_{t+1} = m' |M_t=m, I_t=i)$ for $m, m' \in \mathcal{M}$.
		
		\item The observation density function $\Omega(o |m, i)$ is a distribution over observations $o \in \mathcal{O}$ that occur in state $m \in \mathcal{M}$ after the agent takes action $i$ in the last period, i.e., $\Omega(o |m, i) = \mathbb{P} (O_{t} = o | M_{t}=m, I_{t-1}=i)$.
		
		\item The reward function $R(m, i)$ specifies the immediate reward for each state-action pair $(m,i)$, and we assume the reward function $R: \mathcal{M} \times \mathcal{I} \rightarrow [0, r_{\max}]$ for some constant $r_{\max}>0.$
	\end{itemize}
	
	The following sequence of events occur in order in each period.
	In period $t$, the underlying state transits to $M_{t}$.
	Then the agent observes $O_t \in \mathcal{O}$, whose distribution depends on $M_{t}$ and $I_{t-1}$.
	The agent then chooses an action $I_t \in \mathcal{I}$ and receives reward $R_t$ determined by reward function which depends on the state $M_t$ and the action $I_t$.
	Then the time proceeds to $t+1$ and the state transits to $M_{t+1}$, whose transition probability depends on the action $I_t$.
	
	In the POMDP model, the agent does not observe the state $M_t$, but only the noisy observation $O_t$,
	after which an action is chosen.
	Moreover, since the agent does not know the state $M_t$ at time $t$, it does not know the reward $R_t = R(M_t, I_t)$. 
	Hence, it is typical to assume that the action does not depend on the reward in the literature \citep{krishnamurthy2016partially, cao2007partially}.
	Therefore, the action taken in period $t$, $I_t$, depends on the history up to time $t$, denoted by
	\begin{align}\label{def:history}
		&\mathcal{H}_0 \coloneqq \{I_0\},\\
		&\mathcal{H}_t \coloneqq \{I_0, O_1, \cdots, I_{t-1}, O_t \}, \quad \text{$t\geq 1$}.
	\end{align}
	The agent attempts to optimize the expected cumulative reward over a finite horizon $T$.
	The information structure is illustrated by the graph in
	Figure~\ref{fig:graphical_model}.
	
	\begin{figure}[t]
		\begin{center}
			\includegraphics[width=0.6\linewidth]{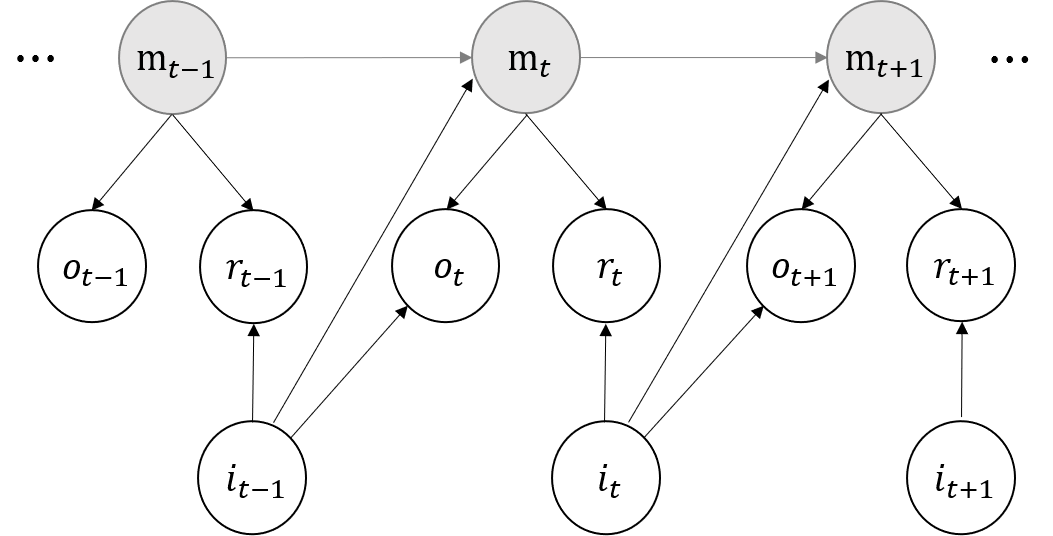}
			\caption{A graph showing the dependence structure of the POMDP.}
			\label{fig:graphical_model}
		\end{center}
	\end{figure}


	
	\subsection{Reformulate POMDP as Belief MDP} \label{subsec:belief_MDP}
	If the model environment in \eqref{eq:pomdp-tuple} is known to the agent,
	then it is well-known (see, e.g., \citealt{krishnamurthy2016partially}) that to maximize the expected reward, the agent can reformulate the POMDP as
	an MDP with a continuous state space.
	The state of the MDP reflects the belief, or the distribution, over the hidden states, and thus it is referred to as the belief MDP.
	More precisely, define an $M$-dimensional vector ${b_t} =(b_t(1),\dots,b_t(M)) \in \mathcal{B} \coloneqq \{b\in \mathbb{R}_+^{M}: \sum_{m=1}^{M}b(m)=1 \}$ as the belief of the underlying state in period $t$:
	\begin{align}\label{def:bt}
		b_0(m)&\coloneqq \mathbb{P}(M_0=m),\\
		b_t(m)&\coloneqq\mathbb{P}(M_t=m|\mathcal{H}_t),\quad t \geq 1.\notag
	\end{align}
	
	Because of the Markovian structure of the belief, we can show (see, e.g., \citealt{krishnamurthy2016partially,puterman2014markov}) that the belief in period $t+1$ can be updated based on the current belief $b_t=b$, the chosen action $I_t=i$ and the observation $O_{t+1}=o$.
	In particular, the updating function
	$H_{\mathcal{P}, \Omega}\colon\mathcal{B}\times\mathcal{I}\times\mathcal{O}\rightarrow\mathcal{B}$ determines
	\begin{align}\label{def:forward-kernel}
		b_{t+1}=H_{\mathcal{P}, \Omega}(b_t,I_t,O_{t+1}).
	\end{align}
	We may omit the dependence on $\mathcal P$ and $\Omega$ if it doesn't cause confusion.
	By Bayes's theorem, we have
	\begin{align}\label{belief_Bayes}
		b_{t+1}(m)&=\frac{\Omega(o|m,i)\sum \limits_{m'\in\mathcal{M}}P^{(i)}(m',m)b_t(m')}{\Prob(o|b,i)},
	\end{align}
	where $\Prob(o|b,i)=\sum \limits_{m''\in\mathcal{M}}\Omega(o|m'',i)\sum\limits_{m'\in\mathcal{M}}P^{(i)}(m',m'')b_t(m')$ is the distribution of the observation under belief $b$ and action $i$.
	
	We next introduce some notations to facilitate the discussion and analysis.
	Define the expected reward conditional on the belief and action
	\begin{align}\label{def:belief-reward-func}
		\bar{R}(b,i)\coloneqq  \sum_{m \in \mathcal{M}}R(m,i)b(m).
	\end{align}
	We can also define the transition kernel of the belief conditional on the action:
	\begin{align}\label{def:belief-trans-kernel}
		\bar{T}(b_{t+1}|b_t,i_t)&:=\Prob(b_{t+1}|b_t,i_t)
		=\sum_{o_{t+1} \in \mathcal{O}} \mathds{1}_{\{H(b_t,i_t,o_{t+1})=b_{t+1}\}} \Prob(o_{t+1}|b_t,i_t).
	\end{align}
	A policy $\mu= (\mu_t)_{t \ge 0}$ for the belief MDP is a mapping from the belief states to actions, i.e., the action chosen by the agent at time $t$ is $I_t = \mu_t (b_t)$.
	Following the literature (see, e.g., \citealt{agrawal2017optimistic}), we define the gain of a policy and the optimal gain.
	\begin{definition}\label{def:gain}
		The gain of a policy $\mu,$ given the initial belief state $b$, is defined as the long-run average reward for the belief MDP over an infinite horizon, given by:
		\begin{align} \label{eq:long-run-average}
			\rho^\mu_b:= \limsup \limits_{T \to \infty}\frac{1}{T}  \mathbb{E} \left[\sum_{t=0}^{T-1} R(M_t, \mu_t) |b_0 =b\right],
		\end{align}
		where the expectation is taken with respect to the interaction sequence  when policy $\mu$ interacts with the belief MDP.
		The optimal gain $\rho^*$ is defined by
		\begin{align}\label{equ:opt average reward}
			\rho^{*}\coloneqq \sup_b \sup_{\mu}\rho_{b}^\mu.
		\end{align}
	\end{definition}

	\subsection{Assumptions}
	Next we provide the technical assumptions for the analysis.
	
	
	\begin{assumption}\label{assum:trans_matrix_min}
		The entries of all transition matrices are bounded away from zero $\epsilon \coloneqq \min\limits_{i \in \mathcal{I}} \min\limits_{m,n \in \mathcal{M}} P^{(i)}(m,n) >0$.
	\end{assumption}

	
	\begin{assumption}\label{assum:reward_density_min}
		$\xi \coloneqq \min\limits_{i \in \mathcal{I}} \min\limits_{o \in \mathcal{O}} \sum\limits_{m \in \mathcal{M}}\Omega(o|m,i) >0$.
	\end{assumption}

	
	Assumptions~\ref{assum:trans_matrix_min} and \ref{assum:reward_density_min} can be strong in general, but they are required by the state-of-art method to bound the belief error caused by the parameter miscalibration (see \cite{de2017consistent} for the HMM setting), which is essential in learning POMDPs. Moreover, the two assumptions provide sufficient conditions to guarantee the existence of the solution to the Bellman optimality equation of the belief MDP and the boundedness of the bias span; See Propositions~\ref{lemma:exist-optimal-policy} and \ref{prop:span-uni-bound}. Note that Assumption~\ref{assum:trans_matrix_min} itself implies that for any fixed $i$, the Markov chain with transition matrix $P^{(i)}$ is geometrically ergodic with
	a unique stationary distribution denoted by $\omega^{(i)}$, and the geometric rate is upper bounded by $1-\epsilon$. See e.g. Theorems 2.7.2 and 2.7.4 in \cite{krishnamurthy2016partially}. This geometric ergodicity, which can hold under weaker assumptions, is needed for spectral estimations of the POMDP model as in \citet{azizzadenesheli2016reinforcement}.

	
	
	



	\begin{assumption}\label{assum:trans_matrix_invert}
		For each $i \in \mathcal{I}$, the transition matrix $P^{(i)}$ is invertible.
	\end{assumption}
	
	

	\begin{assumption}\label{assum:observation_density_lp}
		For all $i\in \mathcal I$, $\Omega(\cdot|1,i),  \cdots, \Omega(\cdot|m,i)$ are linearly independent.
	\end{assumption}

	Assumption~\ref{assum:trans_matrix_invert} and~\ref{assum:observation_density_lp} are required for the finite-sample guarantee of spectral estimators \citep{anandkumar2012method,anandkumar2014tensor}.
	See Section~\ref{sec:spectral} for more details.
	Since our learning algorithm uses the spectral estimator to estimate hidden Markov models, our approach inherits the assumptions.
	
	Before we proceed, we first state a result on the characterization of the optimal gain $\rho^*$ given in Definition~\ref{def:gain} and
	the existence of stationary optimal policies for the belief MDP \eqref{eq:long-run-average} {under the average reward criterion}. Note that in general (without the assumptions), there is no guarantee that a stationary optimal policy would exist for problem \eqref{eq:long-run-average} (see e.g. \citealt{yu2004discretized}). 

	\begin{proposition}\label{lemma:exist-optimal-policy}
		Suppose Assumptions~\ref{assum:trans_matrix_min} and \ref{assum:reward_density_min} hold. There exists a bounded function $v\colon\mathcal{B}\to \R$ and a constant $\rho^* \in \R$ such that the Bellman optimality equation holds for problem \eqref{equ:opt average reward}:
		\begin{align}
			\rho^*+v(b)=\max_{i \in \I}\left[ \bar{R}(b,i)+\int_{\mathcal{B}}v(b')\bar{T}(db'|b,i)\right], \quad \forall b\in\mathcal{B}.\label{equ:Bellman-thm}
		\end{align}
		Moreover, there exists a stationary deterministic optimal policy $\mu^*$ for problem \eqref{equ:opt average reward}, which prescribes an action that maximizes the right side of \eqref{equ:Bellman-thm}. The constant $\rho^*$ is the optimal gain defined in \eqref{equ:opt average reward}.
	\end{proposition}

	\Copy{rev:existence}{\rev{Although Proposition~\ref{lemma:exist-optimal-policy} is necessary for the subsequent analysis, it mainly guarantees the regularity condition and the proof is independent of the algorithmic design.
			To understand its intuition at a high level,
			note that the function $v$ is referred to as the bias function, or the relative value function of the belief state for the undiscounted problem \eqref{eq:long-run-average} (Chapter 8 of \citealt{puterman2014markov}).
			Conditions for the existence are known in the literature; see e.g., \cite{ross1968arbitrary, hsu2006existence}.
			To establish the existence, the bias functions $\{v^{\beta}(b): \beta \in (0,1)\}$ for the infinite-horizon discounted POMDPs with discount factor $\beta$ are studied.
			As $\beta\to 1$, the problems converge to the undiscounted one.
			However, the key condition to ensure the convergence is that the bias functions $\{v^{\beta}\}$ are \emph{uniformly} bounded in $\beta$.
			For this purpose, we use the following ideas.
			First, to bound $v^{\beta}$, it suffices to bound the Lipschitz modulus of the optimal value function for the infinite-horizon discounted problem since the belief state is a probability vector and is bounded.
			Then it reduces to bound the Lipschitz modulus of the optimal value function of the finite-horizon discounted problem by the results in \cite{hinderer2005lipschitz}.
			For finite-horizon problems, we use backward inductions to obtain bounds (uniform in $\beta$) on the Lipschitz modulus of optimal value functions recursively .
			A challenge in the last step is to show the transition kernel of the belief state is a contraction when the number of steps is large, i.e. it has Lipschitz modulus strictly smaller than one. This partly follows from the geometric ergodicity or uniform forgetting property of the belief (see Lemmas~\ref{lemma:bt_exponential_decay} and~\ref{lemma:lip_module_bound} in the appendix).
			Our proof technique yields an explicit upper bound on the bias span in terms of the smallest element of the transition matrices in Assumption~\ref{assum:trans_matrix_min}.
			Such a bound on the bias span is known to be critical in the regret analysis of learning continuous MDPs, see e.g. \cite{lakshmanan2015improved}.}}
	
	We also remark that solving the optimality equation~\eqref{equ:Bellman-thm} and finding the optimal policy for POMDP with average reward criteria in a known environment are computationally challenging due to the continuous belief states. \rev{Various methods have been proposed to compute an approximately optimal policy for belief MDPs or more general continuous-state MDPs with average reward criterion. See, e.g.
		\cite{ormoneit2002kernel, yu2004discretized, yu2008near, saldi2017asymptotic, sharma2020approximate} and the references therein.}
	In this work, we do not focus on this planning problem and assume the access to an optimization oracle that solves the Bellman equation \eqref{equ:Bellman-thm} and returns $\rho^*$ and the optimal stationary policy $\mu^*$.
	

	\subsection{Learning POMDP}\label{sec:learning}
	We consider learning algorithms to learn the POMDP model when some model parameters are unknown.
	In particular, the agent knows the state space $\mathcal{M}$, the action space $\mathcal{I}$, the observation space $\mathcal{O}$, and the reward function $R(m,i)$, but has no knowledge about the underlying hidden state $M_t$, the transition matrices $P^{(i)}$ for all actions and the observation density function $\Omega(o|m,i)$.
	The goal is to design a learning policy to decide which action to take in each period to maximize the expected cumulative reward over $T$ periods even if $T$ is unknown in advance. 
	Note that the setting is slightly different from multi-armed bandits, in which the reward distribution of each arm is unknown.
	In POMDP, it is typical to assume $R(m,i)$ to be a deterministic function and the random noise mainly comes from the observation.
	Moreover, the realized reward is usually not observed or used to determine the action, as mentioned previously.
	Therefore, it is reasonable to set up the environment to learn the parameters related to the observations.
	Our approach can be used to learn the reward function as well, if the historical reward can be observed.
	
	%
	%
	
	For a learning policy $\pi$, the action taken in period $t$, which we denote by $\pi_t$, is adapted to the history $\mathcal{H}_t=\{\pi_0,O^{\pi}_1,...,\pi_{t-1},O^{\pi}_{t}\}$,
	where $O_t^{\pi}$ denotes the observation received under the learning policy $\pi$ in period $t$.
	Note that $\pi_t$ maps the initial belief $b$ and the history $\mathcal H_t$ to an action in period $t$.
	Similar to Definition~\ref{def:gain}, we may define the reward in period $t$ for the policy $\pi$ when the initial belief $b$ as
	\begin{align}\label{def:policy_reward}
		R_t^{\pi}(b)\coloneqq R(M_t, \pi_t).
	\end{align}
	Note that both $M_t$ and $\pi_t$ depend on the initial belief $b$, which we omit in the notation.
	
	To measure the performance of a learning policy, we follow the literature (see, e.g., \citealt{jaksch2010near, ortner2014regret, agrawal2017optimistic}) and set the optimal gain as the benchmark.
	In particular,  we define the total regret of $\pi$ in $T$ periods as
	\begin{align}\label{def:reg}
		\mathcal{R}_T^{\pi}\coloneqq \max_{b}\left\{(T+1)\rho^* - \sum_{t=0}^T R_t^{\pi}(b)\right\}.
	\end{align}
	The objective is to design efficient learning algorithms whose regret grows sublinearly in $T$ with theoretical guarantees. In the sequel, the dependency of $\mathcal{R}_T^{\pi}$ on $\pi$ may be dropped if it is clear from the context.

	\section{The SEEU Learning Algorithm}\label{sec:alg}
	This section describes our learning algorithm for the POMDP, which is referred to as the \emph{Spectral Exploration and Exploitation with Upper Confidence Bound} (SEEU) algorithm.
	We first provide a high-level overview of the algorithm and then elaborate on the details.
	
	To device a learning policy for the POMDP with unknown $\mathcal{P}$ (transition probabilities) and $\Omega$ (observation distributions),
	one needs a procedure to estimate those quantities from the history, i.e., the past actions and observations.
	\citet{anandkumar2012method,anandkumar2014tensor} propose the spectral estimator for the unknown parameters in hidden Markov models (HMMs), with finite-sample theoretical properties.
	It serves as a major component in the SEEU algorithm.
	
	However, the spectral estimator is not directly applicable to ours, because there is no decision making involved in HMMs.
	In a POMDP, the action may depend on past observations and such dependency violates the assumptions of the spectral estimator.
	To address the issue, we divide the horizon $T$ into nested ``exploration'' and ``exploitation'' phases. 
	In the exploration phase, we choose each action successively for a fixed length of periods.
	This transforms the system into an HMM so that we can apply the spectral method to estimate $\mathcal{P}$ and $\Omega$ from the observed actions and observations in that phase.
	In the exploitation phase, based on the confidence region of the estimators obtained from the exploration phase,
	we use a UCB-type policy to implement the optimistic policy (the optimal policy for the best-case estimators in the confidence region) for the POMDP.
	
	The SEEU algorithm is presented in Algorithm~\ref{alg:SEEU}.
	The algorithm proceeds with episodes with increasing length, similar to the UCRL2 algorithm in \cite{jaksch2010near} for learning MDPs.
	Each episode is divided into exploration and exploitation phases.
	The exploration phase lasts $\tau_1I$ periods (Step~\ref{step:explore-tau1}), where $\tau_1$ is a tunable hyperparameter and $I$ is the total number of actions in the action space.
	In this phase, the algorithm chooses each action successively for $\tau_1$ periods.
	In Step~\ref{step:all-samples-exploration} it applies the spectral estimator (Algorithm~\ref{alg:spectral} to be introduced in Section~\ref{sec:spectral}) to (re-)estimate $\mathcal{P}$ and $\Omega$.
	Moreover, it constructs a confidence region based on Proposition~\ref{prop:spectral} with a confidence level $1-\delta_k$, where $\delta_k\coloneqq\delta/k^3$ is a vanishing sequence with $\delta >0$ in episode $k$ (Step~\ref{step:conf-region}).
	The key information to extract from the exploration phase is
	\begin{itemize}
		\item the optimistic POMDP inside the confidence region (Step~\ref{step:optimistic-pomdp});
		\item the updated belief vector according to the new estimators (Step~\ref{step:update-belief}).
	\end{itemize}
	Then the algorithm enters the exploitation phase (Step~\ref{step:exploit}),
	whose length is $\tau_2 \sqrt{k}$ in episode $k$ and $\tau_2$ is another tunable hyperparameter.
	In the exploitation phase, an action is chosen according to the optimal policy associated with the optimistic estimators for $\mathcal{P}$ and $\Omega$ inside the confidence region.
	This is the principle of ``optimisim in the face of uncertainty'' for UCB-type algorithms.
	\begin{algorithm}
		\caption{The SEEU Algorithm}
		\label{alg:SEEU}
		\begin{algorithmic}[1]
			\REQUIRE Precision $\delta$, exploration hyperparameter $\tau_1$,  exploitation hyperparameter $\tau_2$.
			\STATE Initialize: time $T_1=0$, initial belief $b_0$.
			\FOR {$k=1,2,3,\dots$}\label{step:episode}
			\FOR{$t=T_k,T_k+1,\dots,T_k+\tau_1I$}\label{step:explore-tau1}
			\STATE Select each action $\tau_1$ times successively.
			\STATE Observe next observation $o_{t+1}$.
			\ENDFOR
			\STATE\label{step:all-samples-exploration} Use the realized actions and observations in all previous exploration phases $\hat{\mathcal{I}}_k \coloneqq \{i_{T_1:T_1+\tau_1I}, \cdots,i_{T_k:T_k+\tau_1I}
			\}$ and $\hat{\mathcal{O}}_k \coloneqq \{o_{T_1+1:T_1+\tau_1I+1},\cdots,o_{T_k+1:T_k+\tau_1I+1}
			\}$ as input to Algorithm~\ref{alg:spectral} to compute\\
			\centerline{$(\hat{\mathcal{P}}_k, \hat{\Omega}_k) = \textbf{SpectralEstimation}(\hat{\mathcal{I}}_k, \hat{\mathcal{O}}_k)$.}
			\STATE \label{step:conf-region} Compute the confidence region $\mathcal{C}_k(\delta_k)$ centered at $(\hat{\mathcal{P}}_k, \hat{\Omega}_k)$ from \eqref{def:confidence bound} using the confidence level $1-\delta_k=1-\delta/k^3$ such that $\Prob\{(\mathcal{P},\Omega)\in\mathcal{C}_k(\delta_k)\}\geq 1-\delta_k$.
			\STATE\label{step:optimistic-pomdp} Find the optimistic POMDP  in the confidence region ($\rho^*$ given in \eqref{equ:opt average reward} and \eqref{equ:Bellman-thm}):\\
			\centerline{$(\mathcal{P}_k,\Omega_k)=\argmax_{(\mathcal{P},\Omega)\in\mathcal{C}(\delta_k)}\rho^*(\mathcal{P},\Omega)$.}
			\FOR{$t=0,1,\dots,T_k+\tau_1I$}
			\STATE \label{step:update-belief} Update belief $b_t^k$ to $ b_{t+1}^k=H_{\mathcal{P}_k, \Omega_k}(b_t^k,i_t,o_{t+1})$ under the new parameters $(\mathcal{P}_k,\Omega_k)$.
			\ENDFOR
			\FOR{$t=T_k+\tau_1I+1,\dots,T_k+\tau_1I+\tau_2\sqrt{k}$}\label{step:exploit}
			\STATE Execute the optimal policy $\pi^{(k)}$ by solving the Bellman equation \eqref{equ:Bellman-thm} with parameters $(\mathcal{P}_k,\Omega_k)$: $i_t=\pi_t^{(k)}(b_t^k)$.
			\STATE Observe next observation $o_{t+1}$.
			\STATE Update the belief at $t+1$ following
			$b_{t+1}^k=H_{\mathcal{P}_k, \Omega_k}(b_t^k,i_t,o_{t+1})$.
			\ENDFOR
			\STATE $T_{k+1}\gets t+1$
			\ENDFOR
		\end{algorithmic}
	\end{algorithm}
	
	Before getting into the details, we comment on the major difficulties of designing and analyzing such an algorithm.
	To apply the spectral estimator, in the exploration phase the actions are chosen deterministically to ``mimic'' an HMM, as mentioned above.
	This is necessary as the spectral estimator requires fast convergence to a stationary distribution, guaranteed by Assumption~\ref{assum:trans_matrix_min}.
	Moreover, at the first sight, the re-calculation of the belief in Step~\ref{step:update-belief} may deviate significantly from the actual belief using the exact parameters.
	The belief relies on the whole history, and a small error in the estimation may accumulate over $t$ periods and lead to an erroneous calculation.
	We show in Proposition~\ref{prop:lip_bt} that the belief error can actually be well controlled.
	This is important for the algorithm to achieve the sublinear regret.
	
	\Copy{rev:nested}{\rev{We remark that the horizon is divided into nested phases, instead of a single exploration phase followed by a exploitation phase,
			because we do not require the knowledge $T$ in advance.
			If $T$ is known to the agent, then it is indeed true that one can use a single episode (exploration followed by exploitation) to attain the same rate of regret.
			This is often not the case in practice.
	}}

	\subsection{Exploration: Spectral Method} \label{sec:spectral}
	We next zoom in to the exploration phase of a particular episode, in order to show the details of the spectral estimator in Step~\ref{step:all-samples-exploration} and~\ref{step:conf-region} \citep{anandkumar2012method,anandkumar2014tensor,azizzadenesheli2016reinforcement}.
	Suppose the exploration phase lasts from period 0 to $N$, with a fixed action $i_t\equiv i$ and realized observations $\{o_1,o_2,\dots,o_{N+1}\}$ sampled according to the observation density $\Omega$.
	When the action is fixed, the underlying state $M_t$ converges to the steady state geometrically fast due to Assumption~\ref{assum:trans_matrix_min}.
	For the ease of exposition, we assume that the system has reached the steady state at $t=0$.
	In Remark~\ref{rmk:stationary-se}, we discuss how to control the error as the system starts from an arbitrary state distribution.

	For $t \in \{2,\dots,N\}$, we consider three ``views'' $(o_{t-1},o_{t},o_{t+1})$.
	(Here a view is simply a feature of the collected data, a term commonly used in data fusion \citep{zhao2017multi}.
	We stick to the term as in the original description of the spectral estimator.)
	We can see from Figure \ref{fig:graphical_model} that given $m_t$ and $i_t\equiv i$, all the three views are independent.
	Since the system has reached the steady state, the distribution of $(o_{t-1},o_{t},o_{t+1})$ is also stationary.
	The key of the spectral estimator is to express the distribution of $(o_{t-1},o_{t},o_{t+1})$
	as a function of the parameters to learn.
	Then the relevant moments are matched to the samples, which is similar to the spirit of methods of moments.
	
	We represent the views in the vector form for convenience.
	Formally, we encode $o_{t-1}$ into a unit vector $v_{1,t}^{(i)}\in \{0,1\}^{O}$, satisfying $\1_{\{v_{1,t}^{(i)}=\bm{e}_o\}}=\1_{\{O_{t-1}=o\}}$.
	Similarly, $o_{t}$ and $o_{t+1}$ can also be expressed as unit vectors $v_{2,t}^{(i)} \in \{0,1\}^{O}$ and $v_{3,t}^{(i)} \in \{0,1\}^{O}$.
	Define three matrices $A_1^{(i)}, A_2^{(i)} , A_3^{(i)} \in \R^{O \times M} $ for action $i$ such that:
	\begin{align}
		&A_1^{(i)}(o,m)=\Prob(v^{(i)}_{1,t}=\bm{e}_o|m_t=m,i_t=i),\\
		&A_2^{(i)}(o,m)=\Prob(v^{(i)}_{2,t}=\bm{e}_o|m_t=m,i_t=i),\\
		&A_3^{(i)}(o,m)=\Prob(v^{(i)}_{3,t}=\bm{e}_o|m_t=m,i_t=i).\label{multi-view}
	\end{align}
	By stationarity, the distribution of the matrices is independent of $t$.
	We use $\theta_{1,m}^{(i)}, \theta_{2,m}^{(i)}$ and $\theta_{3,m}^{(i)}$ to denote the $m$-th column of $A_1^{(i)}$, $A_2^{(i)}$ and $A_3^{(i)}$, respectively.
	Let $W_{p,q}^{(i)}=\E\left[v_{p,t}^{(i)}\otimes v_{q,t}^{(i)}\right]$ be the correlation matrix between $v_{p,t}^{(i)}$ and $v_{q,t}^{(i)}$, for $p,q\in\{1,2,3\}$.\footnote{For any vectors $v \in \mathbb{R}^{n_1}, u \in \mathbb{R}^{n_2}, w \in \mathbb{R}^{n_3}$, the tensor products are defined as follows: $v \otimes u \in \mathbb{R}^{n_1 \times n_2}$ with $[v \otimes u]_{i,j}=v_i u_j$, and $v \otimes u \otimes w \in \mathbb{R}^{n_1 \times n_2 \times n_3}$ with $[v \otimes u \otimes w]_{i,j,k}=v_i u_j w_k$.}

	The spectral estimator uses the following modified views, which are linear transformations of $v_{1,t}^{(i)}$ and $v_{2,t}^{(i)}$:
	\begin{align}\label{def:modified view}
		\widetilde{v}_{1,t}^{(i)}\coloneqq W_{3,2}^{(i)}(W_{1,2}^{(i)})^\dagger v_{1,t}^{(i)},\quad\quad
		\widetilde{v}_{2,t}^{(i)}\coloneqq W_{3,1}^{(i)}(W_{2,1}^{(i)})^{\dagger}v_{2,t}^{(i)},
	\end{align}
	where $\dagger$ represents the pseudoinverse of a matrix.
	It turns out that the second and third moment of the modified views,
	\begin{align}\label{def:view moments}
		M_2^{(i)}\coloneqq\E\left[\widetilde{v}_{1,t}^{(i)}\otimes\widetilde{v}_{2,t}^{(i)}\right],\quad
		M_3^{(i)}\coloneqq\E\left[\widetilde{v}_{1,t}^{(i)}\otimes\widetilde{v}_{2,t}^{(i)}\otimes v_{3,t}^{(i)}\right],
	\end{align}
	can be compactly represented by the model parameters.
	More precisely, by Theorem 3.6 in \citet{anandkumar2014tensor}, we have the following spectral decomposition: 
	\begin{align}\label{eq:M_tensor_decom}
		M_2^{(i)}=\sum\limits_{m\in\mathcal{M}}\omega^{(i)}(m)\theta_{3,m}^{(i)}\otimes\theta_{3,m}^{(i)},\quad\quad
		M_3^{(i)}=\sum\limits_{m\in\mathcal{M}}\omega^{(i)}(m)\theta_{3,m}^{(i)}\otimes\theta_{3,m}^{(i)}\otimes\theta_{3,m}^{(i)},
	\end{align}
	where we recall that $\omega^{(i)}(m)$ is the state stationary distribution under the policy $i_t\equiv i$ for all $t$. 
	
	With the relationship \eqref{eq:M_tensor_decom}, we can describe the procedures of the spectral estimator.
	Suppose a sample path $\left\{o_t\right\}_{t=1}^{N+1}$ is observed under the policy $i_t\equiv i$.
	It can be translated to $N-1$ samples of $(v_{1,t}^{(i)}, v_{2,t}^{(i)}, v_{3,t}^{(i)})$, $t \in \{2,\dots,N\}$.
	They can be used to construct the sample average of $W_{p,q}^{(i)}$ for $p,q\in\{1,2,3\}$:
	\begin{align}\label{eq:w-hat}
		\hat{W}_{p,q}^{(i)}=\frac{1}{N-1}\sum_{t=2}^N v_{p,t}^{(i)}\otimes v_{q,t}^{(i)}.
	\end{align}
	By \eqref{def:modified view} and \eqref{def:view moments}, we can construct the following estimators:
	\begin{align}
		&\hat{v}_{1,t}^{(i)}=\hat{W}_{3,2}^{(i)}(\hat{W}_{1,2}^{(i)})^{\dagger}v_{1,t}^{(i)},\quad\quad
		&&\hat{v}_{2,t}^{(i)}=\hat{W}_{3,1}^{(i)}(\hat{W}_{2,1}^{(i)})^{\dagger}v_{2,t}^{(i)},\label{eq:y-hat}\\
		&\hat{M}_2^{(i)}=\frac{1}{N-1}\sum_{t=2}^N\hat{v}_{1,t}^{(i)}\otimes \hat{v}_{2,t}^{(i)},\quad\quad
		&&\hat{M}_3^{(i)}=\frac{1}{N-1}\sum_{t=2}^N \hat{v}_{1,t}^{(i)}\otimes \hat{v}_{2,t}^{(i)}\otimes v_{3,t}^{(i)}.\label{eq:M-hat}
	\end{align}
	Plugging $\hat M_2^{(i)}$ and $\hat M_3^{(i)}$ into the left-hand sides of \eqref{eq:M_tensor_decom},
	we can apply the tensor decomposition method \citep{anandkumar2014tensor} to solve $\theta_{3,m}^{(i)}$ from \eqref{eq:M_tensor_decom}, which is denoted as $\hat{\theta}_{3,m}^{(i)}$.
	It can also be shown that $\theta_{1,m}^{(i)}=W_{1,2}^{(i)}(W_{3,2}^{(i)})^ \dagger \theta_{3,m}^{(i)}$ and $\theta_{2,m}^{(i)}=W_{2,1}^{(i)}(W_{3,1}^{(i)})^ \dagger \theta_{3,m}^{(i)}$, which naturally lead to estimators $\hat{\theta}_{1,m}^{(i)}$ and $\hat{\theta}_{2,m}^{(i)}$.
	As a result, the unknown parameters $P^{(i)}$ and $\Omega$ can be estimated according to the following lemma.
	\begin{lemma}\label{lemma:view_to_f_P}
		The unknown transition matrix $P^{(i)}$ and the observation density function $\Omega$ satisfy $\Omega(o|m,i)=A_2^{(i)}(o,m)$ and  $P^{(i)}=\left(\left(A_2^{(i)}\right)^\dagger A_3^{(i)}\right)^{\top}$.
	\end{lemma}
	We remark that Assumption \ref{assum:trans_matrix_invert} and Assumption \ref{assum:observation_density_lp} imply that all three matrices $A_1^{(i)}, A_2^{(i)} , A_3^{(i)} $ are all of full column rank \citep{azizzadenesheli2016reinforcement}, and hence the pseudoinverse $\left(A_2^{(i)}\right)^\dagger$ in Lemma~\ref{lemma:view_to_f_P} is well defined. The subroutine to estimate POMDP estimators is summarized in Algorithm~\ref{alg:spectral}.

	
	\begin{algorithm}[H]
		\caption{The subroutine to estimate POMDP parameters.}
		\label{alg:spectral}
		\begin{algorithmic}[1] 
			\REQUIRE Observed actions $\{i_0,\dots,i_N\}$ and observations $\{o_1,\dots,o_{N+1}\}$
			\ENSURE POMDP parameters $\hat{\mathcal{P}}, \hat{\Omega}$
			\FOR {$i=1,\cdots,I$}
			\STATE Construct $v^{(i)}_{1,t} = o_{t-1}$, $v^{(i)}_{2,t} = o_{t}$ and $v^{(i)}_{3,t}=o_{t+1}$.
			\STATE Compute $\hat{W}_{p,q}^{(i)}, p,q\in\{1,2,3\}$ according to \eqref{eq:w-hat},
			\STATE Compute $\hat v_{1,t}^{(i)}$ and $\hat v_{2,t}^{(i)}$ according to~\eqref{eq:y-hat}.
			\STATE Compute $\hat{M}_2^{(i)}$ and $\hat{M}_3^{(i)}$ according to \eqref{eq:M-hat} .
			\STATE Apply tensor decomposition  (\citet{anandkumar2014tensor}) to compute\\ \centerline{$\hat{A}_3^{(i)}=\textbf{TensorDecomposition}(\hat{M}_2,\hat{M}_3)$.}
			\STATE Compute $\hat{\theta}_{2,m}^{(i)}=\hat{W}_{2,1}^{(i)}(\hat{W}_{3,1}^{(i)})^ \dagger \hat{\theta}_{3,m}^{(i)}$ for each $m\in\mathcal{M}$.
			\STATE Return $\hat{\Omega}(o|m,i)=\hat{A}_2^{(i)}(o,m)$.
			\STATE Return $\hat{P}^{(i)}=\left(\left(\hat{A}_2^{(i)}\right)^\dagger \hat{A}_3^{(i)}\right)^{\top}$.
			\ENDFOR
		\end{algorithmic}
	\end{algorithm}
	
	\begin{remark}\label{rmk:stationary-se}
		The stationary distribution for a fixed action $i_t\equiv i$ is crucial for the spectral estimator $\hat{\Omega}(o|m,i)$ and $\hat{P}^{(i)}$, which allows \eqref{def:view moments} and \eqref{eq:M_tensor_decom} to be independent of $t$.
		In our case, the spectral estimator is applied to a sequence of samples in the exploration phase, which does not start in a steady state.
		This is a similar situation as \cite{azizzadenesheli2016reinforcement}.
		Fortunately, Assumption~\ref{assum:trans_matrix_min} allows fast mixing so that the distribution converges to the stationary distribution at a sufficiently fast rate.
		We can still use Algorithm~\ref{alg:spectral}, which is originally designed for stationary HMMs.
		The theoretical result in Proposition~\ref{prop:spectral} already takes into account the error attributed to mixing.
	\end{remark}
	
	\Copy{rev:spectral}{\rev{Note that Algorithm~\ref{alg:spectral} is directly adapted from the spectral estimator studied in the literature, e.g., \citet{anandkumar2012method,azizzadenesheli2016reinforcement}.}}
	The following result, adapted from \citet{azizzadenesheli2016reinforcement}, provides the confidence regions of the estimators in Algorithm~\ref{alg:spectral}.
	
	\begin{proposition}[Finite-sample guarantee of spectral estimators]\label{prop:spectral}
		Suppose Assumptions \ref{assum:trans_matrix_min}, \ref{assum:trans_matrix_invert} and \ref{assum:observation_density_lp} hold. For any $\delta \in (0,1)$. 
		If for any action $i \in \mathcal{I}$, the number of samples $N^{(i)}$ satisfies $N^{(i)}\geq N_0^{(i)}$ for some $N_0^{(i)}$,
		then with probability $1-\delta$, the estimated $\hat{P}^{(i)}$ and $\hat{\Omega}$ by Algorithm~\ref{alg:spectral} satisfy
		\begin{align}
			\left\|\Omega(\cdot|m,i)-\hat{\Omega}(\cdot|m,i)\right\|_1 &\leq C_1 \sqrt{\frac{\log\left(\frac{6(O^2+O)}{\delta}\right)}{N^{(i)}}},\\
			\left\|P^{(i)}(m,:)- \hat{P}^{(i)}(m,:)\right\|_2&\leq C_2\sqrt{\frac{\log\left(\frac{6(O^2+O)}{\delta}\right)}{N^{(i)}}}.\label{def:confidence bound}
		\end{align}
		for $i \in \mathcal{I}$ and $m \in \mathcal{M}$. Here, $C_1$ and $C_2$ are constants independent of any $N^{(i)}$.
	\end{proposition}
	
	The explicit expressions of constants $N_0^{(i)}, C_1, C_2$ are given in Section~\ref{sec:proof-prop-spectral} in the appendix.
	\begin{remark}
		Note that $\Omega$ and $P^{(i)}$ are identifiable up to a proper permutation of the hidden state labels, because the exact index of the states cannot be recovered.
		In \citet{azizzadenesheli2016reinforcement}, because the reward can be observed, the states associated with the observations and rewards can be correctly identified up to a permutation.
		In this paper, since we follow the standard setup in the POMDP literature and assume unobserved rewards, we have to make sure the inferred states match the indices of the reward function.
		In practice, this issue can be resolved if one can rely on external identifiability conditions to sort and label the states \citep{stephens2000dealing}: for example, in the application of financial economics, 
		state one may represent a bullish market and state two maps to a bearish market. 
		Once the observation density function $\Omega$ is accurately estimated (suppose the observations are macroeconomic indicators), when the sample size $N$ is sufficiently large,
		the agent can naturally construct a mapping to the states.
		This is similar to the assumptions made in \cite{azizzadenesheli2016reinforcement}.
		In the numerical experiment (Section~\ref{sec:numerical}), the algorithm always correctly labels the states.
		We do not explicitly mention the permutation in the statement of Proposition~\ref{prop:spectral} for simplicity, consistent with the literature such as \cite{azizzadenesheli2016reinforcement}.
	\end{remark}

	\subsection{Exploitation: UCB-type Method}
	After the confidence region and the implied coverage probabilities are derived in Proposition~\ref{prop:spectral} at the end of the exploration phase,
	we adopt the principle of ``optimism in the face of uncertainty'' and the associated UCB algorithm \citep{auer2002finite} in the subsequent exploitation phase.
	It is based on the intuition that the upper confidence bound creates a collection of ``plausible'' environments and selects
	the one with the largest optimal gain.
	The UCB algorithm has been successfully applied to reinforcement learning (UCRL) \citep{auer2006logarithmic, jaksch2010near} to control the regret.
	We apply this idea to Step~\ref{step:optimistic-pomdp} of~Algorithm \ref{alg:SEEU}.
	Selecting the optimal action based on the optimistic yet plausible environment helps to balance the exploration and exploitation.

	\subsection{Discussions on the SEEU Algorithm}\label{sec:discussion}
	We discuss a few points related to the implementation of Algorithm~\ref{alg:SEEU}. 

	\textbf{Computational cost of Algorithm~\ref{alg:SEEU}.} For given parameters $(\mathcal{P},\Omega)$, we need to compute the optimal average reward $\rho^*(\mathcal{P},\Omega)$ that depends on the parameters (Step \ref{step:optimistic-pomdp} in Algorithm \ref{alg:SEEU}). Various computational and approximation methods have been proposed in the literature to tackle this planning problem for belief MDPs, which we have already discussed in the introduction. These methods can be applied to our algorithm.
	In addition, we need
	to find out the optimistic POMDP in the confidence region $\mathcal{C}_k(\delta_k)$ with the best average reward (Step \ref{step:optimistic-pomdp} in Algorithm \ref{alg:SEEU}). For low dimensional models,  one can discretize $\mathcal{C}_k(\delta_k)$ into grids and calculate the corresponding optimal average reward $\rho^*$ at each grid point so as to find (approximately) the optimistic model $(\mathcal{P}_k, \Omega_k)$.
	However, in general it is not clear whether there is an efficient computational method to find the optimistic plausible POMDP model in the confidence region when the unknown parameters are high-dimensional. This issue is also present in other recent studies on learning continuous-state MDPs with the upper confidence bound approach, see e.g. \cite{lakshmanan2015improved} for a discussion. In our regret analysis below, we do not take into account the approximation errors arising from the computational aspects discussed here.
	We point out that the main contribution of this paper is not \textit{computational}.
	Rather, it is to develop a theoretical framework for learning model-based POMDPs.
	To implement the algorithm efficiently remains an intriguing  future direction.
	
	\textbf{Dependence on the unknown parameters.}
	Algorithm~\ref{alg:SEEU} requires some information about the unknown Markov chain to compute the confidence bounds.
	In particular, when computing the confidence region in Step~\ref{step:conf-region} of Algorithm~\ref{alg:SEEU}, the agent needs the information of the constants $C_1$ and $C_2$ in Proposition~\ref{prop:spectral}.
	Although $C_1$ and $C_2$ do not depend on the parameters to learn directly, such as the transition matrices, they do depend on a few ``primitives'' that are hard to know, for example, the mixing rate of the underlying Markov chain when the action is fixed.
	See Section~\ref{sec:proof-prop-spectral} in the appendix for more details.
	Still, we would argue that it is relatively easy to acquire such information and this setup is innocuous:
	we only need upper bounds for $C_1$ and $C_2$ for the theoretical guarantee and not exact values.
	Therefore, a rough and conservative estimate would be sufficient.
	Such dependence on some unknown parameters is common in learning problems: the parameter of the sub-Gaussian noise in multi-armed bandits is usually assumed to be known; the confidence bound in \citet{azizzadenesheli2016reinforcement} is constructed based on similar information of the underlying Markov chain; \cite{ortner2014regret,lakshmanan2015improved} require the knowledge of the H\"older constant for rewards and transition probabilities.
	A common remedy is to dedicate the beginning of the horizon to estimate the unknown parameters, which typically doesn't increase the rate of the regret.
	Alternatively, $C_1$ and $C_2$ can be replaced by parameters that are tuned by hand.
	See Remark 3 of \cite{azizzadenesheli2016reinforcement} for a discussion on this issue.
	
	\textbf{The ETC algorithm.}
	As an alternative to the SEEU algorithm, the ETC (Explore-then-Commit) algorithm can also be applied to POMDPs.
	The structure of the ETC algorithm is similar to the SEEU algorithm. The main difference is that in each exploitation phase, the ETC algorithm uses the point estimator of the POMDP parameters directly while the SEEU algorithm uses the optimistic estimator in the confidence region.
	The detailed steps of the algorithm are summarized in Algorithm~\ref{alg:etc} in the appendix. 
	The ETC algorithm is conceptually simpler than the SEEU algorithm, because it doesn't need to calibrate the size of the confidence region and simply uses the point estimators.
	However, the regret analysis requires a different set of techniques, in particular, the sensitivity of POMDPs to its parameters.
	We show that it achieves the same rate of regret as the SEEU algorithm in Section~\ref{sec:etc} in the appendix.

	\section{Regret Analysis}\label{sec:regret}
	In this section, we state our main results, the upper bound for the regret of Algorithm~\ref{alg:SEEU}
	in high probability and expectation.
	We first show a uniform bound on the span of $v_k$, where $v_k$ is the bias function satisfying the Bellman equation \eqref{equ:Bellman-thm} for the optimistic belief MDP in episode $k$ of Algorithm \ref{alg:SEEU}.
	Such a bound is known to be critical in the regret analysis of learning continuous MDPs.
	

	\begin{proposition}[Uniform bound for the span of the bias function]\label{prop:span-uni-bound}
		Suppose Assumption \ref{assum:trans_matrix_min} and Assumption \ref{assum:reward_density_min} hold, and $(\rho^*_k, v_k)$ satisfies the Bellman equation \eqref{equ:Bellman-thm}. Fix the hyperparameter $\tau_1$ in Algorithm~\ref{alg:SEEU} to be sufficiently large.
		Then there exists a constant $D$ such that $\text{span}(v_k) :=\max_{b \in \mathcal{B}}v_k(b)-\min_{b\in\mathcal{B}}v_k(b)\le D$ for all $k$, where
		\begin{align}
			D\coloneqq\frac{8r_{\max}\left( \frac{2}{(1-\bar \alpha)^2}+(1+ \bar \alpha) \log_{\bar \alpha} \frac{1-\bar \alpha}{8}\right)}{1-\bar \alpha},\label{span-bound-constant}
		\end{align}
		and $\bar \alpha=\frac{1- \epsilon}{1 - \epsilon/2} \in (0,1)$ with   $\epsilon=\min\limits_{i \in \mathcal{I}} \min\limits_{m,n \in \mathcal{M}}P^{(i)}(m,n)>0$.
	\end{proposition}
	
	Next, recall that in Algorithm \ref{alg:SEEU}, we re-compute the belief after re-estimating the parameters in each exploration phase.
	If a small error in the parameter estimation may propagate over time and cause the belief to deviate from the true value, then the algorithm cannot perform well in the exploitation phase.
	In the next result (Proposition~\ref{prop:lip_bt}), we show guarantees that the error doesn't accumulate and as a result, the regret incurred in the exploitation phase is proportional to the estimation error.
	
	\begin{proposition}[Controlling the belief error]\label{prop:lip_bt}
		Suppose Assumption \ref{assum:trans_matrix_min} and Assumption \ref{assum:reward_density_min} hold.
		Given the estimators $((\hat{P}^{(i)})_{i},\hat{\Omega})$ of the true model parameters $( (P^{(i)})_{i},\Omega)$, for an arbitrary reward-action sequence $\{i_{0:t-1},o_{1:t}\}_{t\geq 1}$, let $b_t$ and $\hat{b}_t$ be the corresponding beliefs in period $t$ computed from the same initial belief $b_0$ under $( (P^{(i)})_{i},\Omega)$ and $((\hat{P}^{(i)})_{i},\hat{\Omega})$, respectively. Then there exist constants $L_1,L_2$ such that
		\begin{align}
			\Vert b_t -\hat{b}_t \Vert_1\leq L_1 \max\limits_{m \in \mathcal{M}, i\in \mathcal{I}}\left\|\Omega(\cdot|m,i)-\hat{\Omega}(\cdot|m,i)\right\|_1+L_2 \max\limits_{m \in \mathcal{M}, i\in \mathcal{I}} \left\| P^{(i)}(m,:)-\hat{P}^{(i)}(m,:) \right\|_2,
		\end{align}
		where $L_1=\frac{4(1-\epsilon)^2}{\epsilon^2 \xi}$ and
		$L_2=\frac{4(1-\epsilon)^2}{\epsilon^3}$.
	\end{proposition}

	With Propositions~\ref{prop:span-uni-bound} and~\ref{prop:lip_bt}, we are now ready to state our main result about the high-probability regret bound for our algorithm.

	\begin{theorem}\label{thm:upper_bound}
		Fix the hyperparameter $\tau_1$ in Algorithm~\ref{alg:SEEU} to be sufficiently large. Suppose Assumptions 1 to 4 hold.
		There exist constants $T_0$ such that for $T>T_0$, with probability at least $1-\frac{7}{2}\delta$,
		the regret of Algorithm~\ref{alg:SEEU} satisfies
		\begin{align}
			\mathcal{R}_T& \leq CT^{2/3}\sqrt{\log\left(\frac{9(O+1)}{\delta}T\right)}+T_0\rho^*,
		\end{align}
		where $\rho^* \le r_{\max},$ and
		\begin{align}
			C=&3\sqrt{2}\left[\left(\frac{D}{2}+\frac{D}{2}L_1 + r_{\max} L_1\right)C_1+\left(\frac{D\sqrt{M}}{2}+\frac{D}{2}L_2+r_{\max} L_2 \right)C_2 \right]\tau_1^{-1/2}\tau_2^{1/3}\\
			&\quad\quad+3(\tau_1I\rho^*+D)\tau_2^{-2/3}+D \sqrt{2 \log\left(\frac{1}{\delta}\right)}+\sqrt{2r_{\max}\log \left(\frac{1}{\delta}\right)}, \label{const: C}
		\end{align}
		with $C_1, C_2$ given in Proposition~\ref{prop:spectral}, $D$ given in Proposition~\ref{prop:span-uni-bound}, and $L_1, L_2$ given in Proposition~\ref{prop:lip_bt}.
	\end{theorem}
	

	We briefly discuss the dependency of $C$ on the model primitives.
	The dependence is square-root in $M$ (similar to the learning of MDPs in \citealt{jaksch2010near}) and linear in $I$.
	In contrast, the regret of MAB typically scales in $\sqrt{I}$. This is because only one arm emerges optimal.
	In our setting, all actions may be optimal depending on the belief of the underlying state, and thus their parameters need to be learned equally accurately.
	The dependency on $C_1$ and $C_2$ is directly inherited from the confidence bounds in Proposition~\ref{prop:spectral} (see also \citealt{azizzadenesheli2016reinforcement}).
	In addition, $C$ depends on $L_1$ and $L_2$ which arise from controlling the propagated error when updating the belief of the hidden state (Proposition~\ref{prop:lip_bt}; see also \citealt{de2017consistent}). Finally, $C$ depends on the bound $D$ of the bias span (similar to the diameter of MDP in UCRL2 in \citealt{jaksch2010near}) in Proposition~\ref{prop:span-uni-bound}. The constant
	$C$ may not be tight, and its dependence on some parameters may be just an artefact of our proof,
	but it is the best bound we can obtain.

	

	Since $\mathcal{R}_T \le (T+1)\rho^* $ by the definition \eqref{def:reg}, we can choose $\delta=\frac{9(O+1)}{T+1}$ in Theorem~\ref{thm:upper_bound},
	and obtain
	\begin{align}
		\E[\mathcal{R}_T]&\leq \left(CT^{2/3}\sqrt{\log\left(\frac{9(O+1)}{\delta}T\right)}+T_0\rho^*\right)\left(1-\frac{7}{2}\delta\right)+(T+1)\rho^* \cdot \frac{7}{2}\delta    \\
		&\leq CT^{2/3}\sqrt{2\log T}+T_0\rho^*+\frac{63(O+1)}{2}\rho^*.
	\end{align}
	We summarise this bound for the expected regret in the following result.

	\begin{theorem}\label{thm:expected_upper_bound}
		Under the same setting as in Theorem~\ref{thm:upper_bound}, the regret of Algorithm~\ref{alg:SEEU} satisfies
		\begin{align}
			\E[\mathcal{R}_T]& \leq CT^{2/3}\sqrt{2\log T} + (T_0+32(O+1))\rho^*,
		\end{align}
		where constants $C$ and $T_0$ are given in Theorem \ref{thm:upper_bound}.
	\end{theorem}
	
	\begin{remark}[Lower bound of the regret]
		The typical optimal regret bound for online learning is $O(T^{1/2})$.
		The gap is probably caused by the split of exploration/exploitation phases in our algorithm, which resembles the $O(T^{2/3})$ regret for explore-then-commit algorithms in classic multi-armed bandit problems (see Chapter 6 in \citealt{lattimore2020bandit}).
		We cannot integrate the two phases because of the following barrier: the spectral estimator cannot use samples generated from the belief-based policy due to the history dependency.
		This is also why \citet{azizzadenesheli2016reinforcement} focus on memoryless policies in POMDPs to apply the spectral estimator.
		In some simpler settings such as linear bandit, the exploration-exploitation interleaving approach can lead to $O(\sqrt{T})$ regret by adaptively varying the length of the phases, see e.g. \cite{rusmevichientong2010linearly}.
		The reason for this difference in regret lies in the structure of the problem.
		In \citet{rusmevichientong2010linearly}, the linear parametrization and smoothness assumption on the set of actions give rise to an instantaneous regret that is proportional to $\|\text{estimation error}\|_2^2$ in the exploitation phase, while for us it is proportional to $\|\text{estimation error}\|_2$.
	\end{remark}

	\section{Numerical Experiments} \label{sec:numerical} 
	In this section, we present proof-of-concept experiments to demonstrate the performance of the SEEU algorithm.
	Note that for large-scale POMDPs, it is computationally expensive even to solve the oracle, i.e., finding the optimal policy to maximize the long-run average reward in a known environment.
	Therefore, we focus on small-scale experiments, following some recent literature on reinforcement learning for POMDPs \citep{azizzadenesheli2016reinforcement, igl2018deep}. 
	
	As a representative example, we consider a POMDP model \eqref{eq:pomdp-tuple} with 2 hidden states, 2 actions, and 2 possible observations where the model parameters given as follows. \vspace{2mm}
	\begin{itemize}
		\item The transition matrices $P^{(i)}$ for action $i=1, 2,$ are
		$P^{(1)}= \left(
		\begin{matrix}
			0.2 & 0.8 \\
			0.9 & 0.1
		\end{matrix}
		\right) $ and
		$P^{(2)}= \left(
		\begin{matrix}
			0.6 & 0.4 \\
			0.3 & 0.7
		\end{matrix}
		\right) $; \vspace{2mm}
		\item The observation densities are
		$\Omega^{(1)}= \left(
		\begin{matrix}
			0.7 & 0.3 \\
			0.4 & 0.6
		\end{matrix}
		\right) $ and
		$\Omega^{(2)}= \left(
		\begin{matrix}
			0.2 & 0.8 \\
			0.9 & 0.1
		\end{matrix}
		\right) $ where $\Omega^{(i)}(m,o) \coloneqq \Omega(o|m,i)$; \vspace{2mm}
		\item The reward function is
		$R= \left(
		\begin{matrix}
			1 & 4 \\
			3 & 2
		\end{matrix}
		\right) $, where $R(m,i)$ denotes the reward for the pair $(m,i).$
		\vspace{2mm}
	\end{itemize}
	
	We compare the regret of our algorithm with two other benchmarks.
	First, we implement the ETC-type algorithm mentioned in Section~\ref{sec:discussion}.
	Second, we also implement the optimal memoryless policy for POMDPs with known parameters discussed in \cite{azizzadenesheli2016reinforcement}. A memoryless policy maps the observation in the current period to an action.
	In Figure \ref{fig:numerical-regret}, we plot the average regret versus $T$ in the log-log scale. Due to computational cost of the search for the optimistic parameters in the confidence region for the SEEU algorithm,
	we run 100 replications of three algorithms. The relative standard error (the standard error divided by the estimated mean) of all three methods are less than 10\%. For SEEU and ETC algorithms, we choose the hyperparameters $\tau_1=5000$ and $\tau_2 =10000.$ 
	The choices of these parameters do not affect the order of the regret as shown in our theoretical results. 
	We can observe from Figure \ref{fig:numerical-regret} that the memoryless policy suffers linear regret.
	This is expected because the optimal policy for such a POMDP is a belief-based policy which relies on the historical observations and memoryless policies are generally suboptimal. On the other hand,
	both the SEEU and the ETC algorithm have sublinear growth of regret, where the slopes of the corresponding curves are both close to $2/3$.
	The similar trends between SEEU and ETC is due to fact that they both share the same nested structure of exploration and exploitation phases. 
	
	\begin{figure}
		\centering
		\includegraphics[width=10cm, height=7cm]{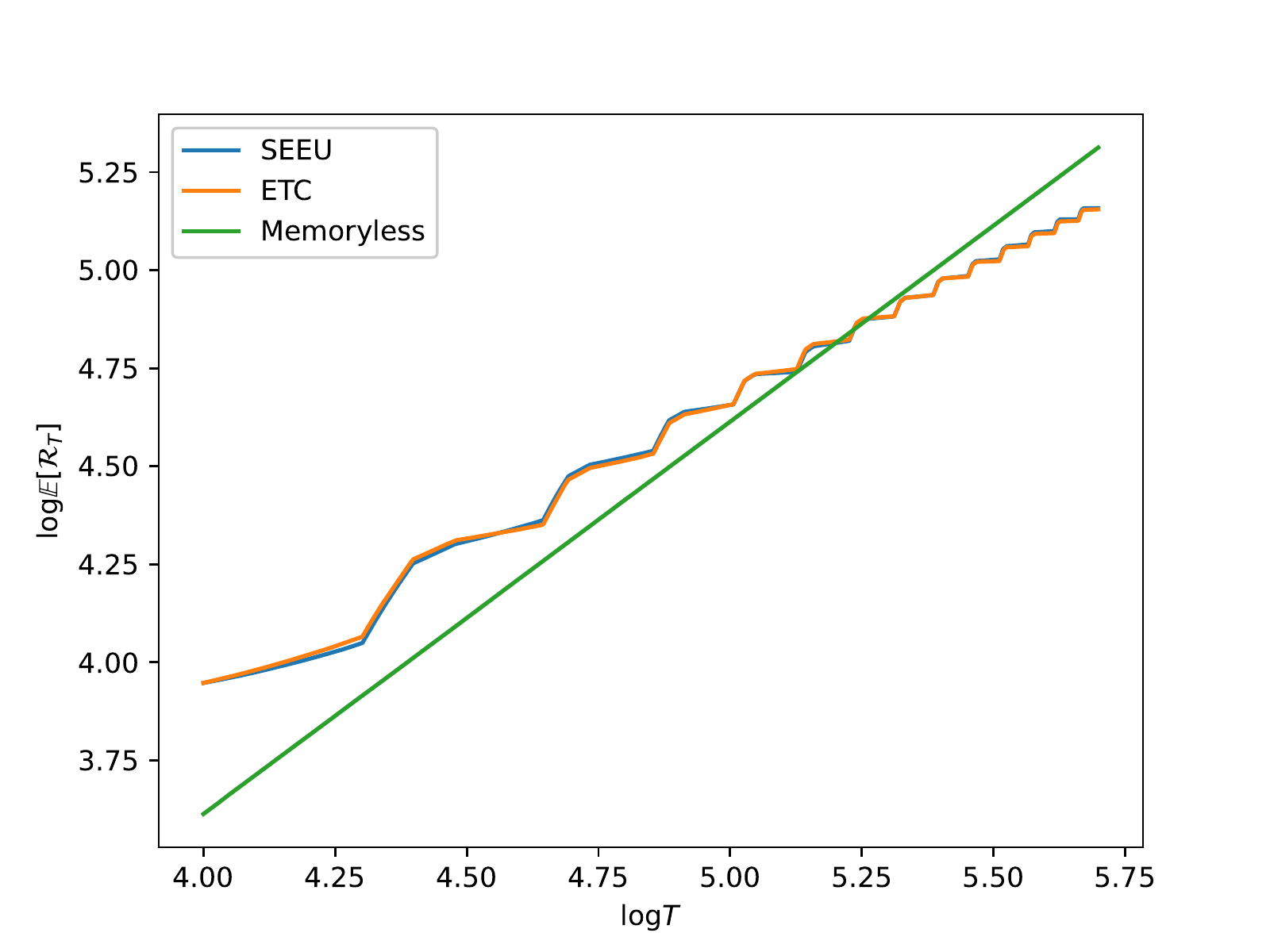}
		\caption{Regret comparison of different algorithms}
		\label{fig:numerical-regret}
	\end{figure}

	\section{Conclusion and Future Work} \label{sec:conclusion}
	This paper studies learning of POMDPs in an unknown environment under the average-reward criterion.
	We develop a learning algorithm that integrates spectral estimators for hidden Markov Models and upper confidence methods from online learning.
	We also establish a regret bound of order of $O(T^{2/3}\sqrt{\log T})$ for the learning algorithm. 
	
	There are two research directions based on the work that is worth exploring.
	\rev{First,
		it is not clear if other algorithms can attain regret of order $\sqrt{T}$ or the regret lower bound is $T^{2/3}$. A related open problem is whether spectral methods can be applied to samples generated from adaptive policies, so that exploration and exploitation can be integrated to improve the theoretical regret bound.}
	We hope to address both problems in future studies.

	\ACKNOWLEDGMENT{We thank the department editor, the senior editor and two referees for many constructive comments which have led
		to an improved version of the paper. The research of Ningyuan Chen is supported by NSERC Discovery Grants Program RGPIN-2020-04038. The research of Xuefeng Gao is supported by Hong
		Kong RGC GRF (No. 14201520 and No. 14201421).}

	
%
	\bibliographystyle{informs2014} 
	\bibliography{ref}
	


	\ECSwitch 
	
	\ECHead{E-Companion for ``Sublinear regret for learning POMDPs"}
	
	\section{Two preliminary lemmas}
	This section states two preliminary lemmas that will be used later.
	The first result states that the Bayesian filter in \eqref{def:forward-kernel}, i.e. the belief, forgets its initial condition geometrically fast for all possible action-observation sequences under our assumptions. In the context of HMM, this is often referred to as the uniform forgetting property, see e.g. Chapter 4.3 in \cite{Cappe2006}.
	
	\begin{lemma}\label{lemma:bt_exponential_decay}
		Suppose Assumptions~\ref{assum:trans_matrix_min} and \ref{assum:reward_density_min} hold. Let $b_0$ and $b'_0$ be two different initial beliefs, then for $t\geq 1$ and an arbitrary action-observation sequence $\{i_{0:t-1}, o_{1:t}\}$, let $b_t=H_{\mathcal{P}, \Omega}^{(t)}(b_0, i_{0:t-1}, o_{1:t})$ and $b_t'=H_{\mathcal{P},\Omega}^{(t)}(b'_0,i_{0:t-1}, o_{1:t})$ be the corresponding beliefs at time $t$ computed according to the same forward kernel $H_{\mathcal{P}, \Omega}$, we have
		\begin{align}
			\Vert b_t-b_t'\Vert _1\leq C_3\alpha^{t}\Vert b_0-b_0'\Vert_1,
		\end{align}
		where $C_3=\frac{2(1-\epsilon)}{\epsilon}, \alpha=1-\frac{\epsilon}{1-\epsilon}$, and $\epsilon=\min\limits_{i \in \mathcal{I}} \min\limits_{m,n \in \mathcal{M}}P^{(i)}(m,n)$.
	\end{lemma}
	
	Lemma~\ref{lemma:bt_exponential_decay} will be used in the proofs of Propositions~\ref{lemma:exist-optimal-policy}, \ref{prop:span-uni-bound} and \ref{prop:lip_bt}.
	Its proof largely follows the arguments in the proof of Theorem 3.7.1 in \cite{krishnamurthy2016partially} for the HMM setting, with minor changes to take into account of the action sequence. Since Assumptions~\ref{assum:trans_matrix_min} and \ref{assum:reward_density_min} provide strong mixing conditions uniformly over the actions, the result readily follows and so we omit the details of the proof.


	The second result, taken from Lemma 4.3.3 in \cite{Cappe2006}, will be used in the proof of Proposition \ref{prop:lip_bt}.
	\begin{lemma}\label{lemma:measure-holder}
		For any two probability measures $\nu^*$ and $\hat{\nu}$ supported on the space $\mathcal{M}=\{1, \ldots, M\}$, define $\nu^*(h)=\int h d \nu^*$ and $ \hat{\nu}(h)=\int h d \hat{\nu}$. If there exists some constant $C$ such that $|\nu^*(h)-\hat{\nu}(h)| \leq C ||h||_{\infty}$ for any bounded function $h$ on $\mathcal{M}$, then $||\nu^*-\hat{\nu}||_1 \leq C$.
	\end{lemma}

	\section{Proofs of Propositions~\ref{lemma:exist-optimal-policy} and \ref{prop:span-uni-bound}}\label{sec:proof _of_exist_optimal_policy}
	\subsection{Proof of Proposition~\ref{lemma:exist-optimal-policy}}
	Note the belief MDP (under the average-reward criterion) we consider has continuous belief state space $\mathcal{B}$, finite action space $\mathcal{I}$, and bounded one step reward function $\bar{R}(b,i)$. We apply Theorem 7 in \cite{hsu2006existence} to prove the existence of a bounded function $v\colon\mathcal{B}\to \R$ and a constant $\rho^*$ that satisfy the Bellman optimality equation \eqref{equ:Bellman-thm}. The existence of a stationary deterministic policy then follows from Theorem 11 in \cite{hsu2006existence}.
	
	We first introduce the discounted problem and a few notations.
	For any policy $\mu$, and discount factor $\beta \in (0,1)$, consider maximizing
	\begin{align}\label{eq:inf-horizon-discount}
		V_{\beta}^\mu(b):= \sum_{t=0}^\infty \beta^t  \mathbb{E}_{\mu} \left[\bar{R}(b_t, i_t) |b_0 =b\right],
	\end{align}
	where the one-step reward $\bar{R}$ is given in \eqref{def:belief-reward-func} and it is bounded. Since the action space $\mathcal{I}$ is finite, one can show (see e.g. \cite{blackwell1965discounted}) that
	there is a stationary deterministic policy $\mu_{\beta}$ that is optimal for the problem \eqref{eq:inf-horizon-discount}, and we denote the corresponding optimal value function by $V_{\beta}^*(b) = \max_{\mu} V_{\beta}^\mu(b)$.
	Pick any fixed belief state $s \in \mathcal{B}$. Define the bias function $v_{\beta}(b)$ for the infinite-horizon discounted problem \eqref{eq:inf-horizon-discount}:
	\begin{align}\label{def:discount bias func}
		v_{\beta}(b)\coloneqq V^*_{\beta}(b)-V^*_{\beta}(s), \quad b \in \mathcal{B}.
	\end{align}
	
	To apply Theorem 7 in \cite{hsu2006existence}, we need to verify that $\{v_{\beta}: \beta \in (0,1)\}$ is a uniformly bounded family of functions.
	We show below that $|v_{\beta}(b)| \leq L$ for a constant $L$ that is independent of $b$ and $\beta$. To this end,
	we first introduce $\ell_1$ distance to the belief space $\mathcal{B}$: $\rho_{\mathcal{B}}(b,b'):=\|b-b'\|_1$. For any function $h:\mathcal{B} \mapsto \R$, define the Lipschitz module of a function $h$ by
	\begin{align}\label{def:lip-module}
		l_{\rho_{\mathcal{B}}}(h)\coloneqq\sup\limits_{b\neq b'} \frac{|h(b)-h(b')|}{\rho_{\mathcal{B}}(b,b')}.
	\end{align}
	The function $h$ is Lipschitz continuous if $l_{\rho_{\mathcal{B}}}(h) <\infty$. Then we can infer from \eqref{def:discount bias func} and \eqref{def:lip-module} that
	\begin{align}\label{inequ:span-bound}
		|v_{\beta}(b)| \leq l_{\rho_{\mathcal{B}}}(V^*_{\beta}) \cdot ||b-s||_1 \le  2 \cdot l_{\rho_{\mathcal{B}}}(V^*_{\beta}), \quad \text{for any $b \in \mathcal{B}.$}
	\end{align}
	Hence, it suffices to bound the Lipschitz module of $V^*_{\beta},$ the optimal value function of the infinite-horizon discounted problem \eqref{eq:inf-horizon-discount} with discount factor $\beta \in (0,1).$ We use an approach based on \cite{hinderer2005lipschitz}, which provides general tools for proving Lipschitz continuity of value functions in MDP's with general state spaces. To bound $l_{\rho_{\mathcal{B}}}(V^*_{\beta})$, it reduces to bound $l_{\rho_{\mathcal{B}}}(V^*_{n,\beta})$, where $V^*_{n,\beta}$ is the optimal value function for the $n-$horizon discounted problem:
	\begin{align}
		V_{n, \beta}^\mu(b):= \sum_{t=0}^{n-1} \beta^t  \mathbb{E}_{\mu} \left[\bar{R}(b_t, i_t) |b_0 =b\right],
	\end{align}
	where $\mu$ is an admissible policy.
	This is because $\lim_{n \rightarrow\infty}V_{n,\beta}^*(b) = V_{\beta}^*(b) \le \frac{r_{\max}}{1-\beta}$ (Proposition 1 in \cite[Chapter 6]{bertsekas1976dynamic}), and then Lemma 2.1(e) in \cite{hinderer2005lipschitz} implies that
	\begin{align}\label{lip-module-average}
		l_{\rho_{\mathcal{B}}}(V_{\beta}^*)\leq \liminf\limits_{n\to\infty}l_{\rho_{\mathcal{B}}}(V_{n,\beta}^*).
	\end{align}
	We next bound $l_{\rho_{\mathcal{B}}}(V_{n,\beta}^*).$ The strategy is to apply the results including Lemmas 3.2 and 3.4 in \cite{hinderer2005lipschitz}, but it requires a new analysis to verify the conditions there. To proceed,
	standard dynamic programming theory states that $V_{n,\beta}^*(b) = J_{0}(b)$, and $J_{0}(b)$ can be computed by the backward recursion:
	\begin{align}
		J_n(b_n) &= \bar{R}(b_n, i_n), \label{equ:Bellman-thm-finite-discount1} \\
		J_{t}(b_t) &= \max_{i_t \in \mathcal{I}} \left\{\bar{R}(b_t, i_t) + \beta \int_{\mathcal{B}} J_{t+1}(b_{t+1}) \bar{T}(db_{t+1}| b_t, i_t)\right\}, \quad 0 \leq t<n, \label{equ:Bellman-thm-finite-discount2}
	\end{align}
	where $\bar{T}$ is the (action-dependent) one-step transition law of the belief state, and $J_{t}(b_t)$ are finite for each $t.$
	More generally, for a given sequence of actions $i_{0:n-1},$ the $n$-step transition kernel for the belief state is defined by
	\begin{align}\label{def:n-trans-ker}
		\bar{T}^{(n)}(\textbf{A}|b,i_{0:n-1}) \coloneqq \Prob(b_{n}\in \textbf{A} |b_0=b,i_{0:n-1}), \quad \textbf{A} \subset \mathcal{B}.
	\end{align}
	To use the results in \cite{hinderer2005lipschitz}, we need to study the Lipschitz property of this multi-step transition kernel as we will see later. Following \cite{hinderer2005lipschitz},
	we introduce the Lipschitz module for a transition kernel $\phi(b, db')$ on belief states.
	Let $K_{\rho_{\mathcal{B}}}(\nu,\theta)$ be the Kantorovich metric of two probability measures $\nu,\theta$ defined on $\mathcal{B}$:
	\begin{align}\label{def:K-metric}
		K_{\rho_{\mathcal{B}}}(\nu,\theta)\coloneqq\sup_f\left\{\left|\int_{\mathcal{B}} f(b)\nu(db)-\int_{\mathcal{B}} f(b)\theta(db)\right|,f\in\text{Lip}_1(\rho_{\mathcal{B}})\right\},
	\end{align}
	where $\text{Lip}_1(\rho_{\mathcal{B}})$ is the set of functions on $\mathcal{B}$ with Lipschitz module $l_{\rho_{\mathcal{B}}}(f)\leq1$.
	Then the Lipschitz module of the transition kernel $l_{\rho_{\mathcal{B}}}(\phi)$ is defined as:
	\begin{align}\label{def:lip-module-transi-kernel}
		l_{\rho_{\mathcal{B}}}(\phi)\coloneqq\sup_{b^1\neq b^2}\frac{K_{\rho_{\mathcal{B}}}(\phi(b^1, db'),\phi(b^2,db'))}{\rho_{\mathcal{B}}(b^1,b^2)}.
	\end{align}
	The transition kernel $\phi$ is called Lipschitz continuous if $l_{\rho_{\mathcal{B}}}(\phi) < \infty$. To bound $l_{\rho_{\mathcal{B}}}(V_{n,\beta}^*)$ and to apply results in \cite{hinderer2005lipschitz}, the key technical result we need is the following lemma. We defer its proof to the end of this section. Recall that $\epsilon=\min\limits_{i \in \mathcal{I}} \min\limits_{m,n \in \mathcal{M}}P^{(i)}(m,n)>0$.
	\begin{lemma}\label{lemma:lip_module_bound}
		For $1 \leq n < \infty$, the $n$-step belief state transition kernel $\bar{T}^{(n)}(\cdot|b,i_{0:n-1})$ in \eqref{def:n-trans-ker} is uniformly Lipschitz in $i_{0:n-1}$, and the Lipschitz module is bounded as follows:
		\begin{align}
			l_{\rho_{\mathcal{B}}}(\bar{T}^{(n)}) \le  C_3\alpha^{n}+C_4,
		\end{align}
		where $C_3=\frac{2}{1 - \alpha}$ and $C_4= \frac{1}{2} + \frac{\alpha}{2}$ with $\alpha=1-\frac{\epsilon}{1-\epsilon} \in (0, 1)$. As a consequence,
		there exist constants $n_0 \in \mathbb{Z}^+$ and $\gamma<1$ such that
		$l_{\rho_{\mathcal{B}}}(\bar{T}^{(n_0)})<\gamma$ for any $i_{0:n-1}$. Here, we can take $n_0=\lceil \log_{\alpha}\frac{1-C_4}{2C_3} \rceil$, and $\gamma=\frac{1}{2}(1+C_4) = \frac{3 + \alpha}{4}$.
	\end{lemma}
	
	With Lemma~\ref{lemma:lip_module_bound}, we are now ready to bound $l_{\rho_{\mathcal{B}}}(V_{n,\beta}^*)$. Consider $n=k n_0$ for some positive integer $k$. We can infer from the value iteration in \eqref{equ:Bellman-thm-finite-discount2} that
	\begin{align}
		J_{t}(b_{t})=\sup_{i_{t:t+n_0-1}}&\Big\{\sum_{\tau=0}^{n_0-1}\beta^\tau\int_{\mathcal{B}}\bar{R}(b_{t+\tau},i_{t+\tau})\bar{T}^{(\tau)}(db_{t+\tau}|b_{t},i_{t:t+\tau-1})\\
		&\quad+\beta^{n_0}\int_{\mathcal{B}}J_{t+n_0}(b_{t+n_0})\bar{T}^{(n_0)}(db_{t+n_0}|b_{t},i_{t:t+n_0-1})\Big\}, \quad 0 \leq t \leq n-n_0. \label{equ:n-Bellman-optimality}
	\end{align}
	It is easy to verify that $J_n(b_n)=\bar{R}(b_n, i_n)$ is uniformly Lipschitz in $i_n$ with Lipschitz module $r_{\max}$. We
	can then infer from Lemma~\ref{lemma:lip_module_bound}, \eqref{equ:n-Bellman-optimality}, and Lemmas 3.2 and 3.4 in \citep{hinderer2005lipschitz} that
	\begin{align}
		l_{\rho_{\mathcal{B}}}(J_{t})&\leq r_{\max}\cdot \sum_{\tau=0}^{n_0-1}\beta^{\tau} l_{\rho_{\mathcal{B}}}^{{\mathcal{I}}^{\tau}}(\bar{T}^{(\tau)})+\beta^{n_0} \cdot l_{\rho_{\mathcal{B}}}^{{\mathcal{I}}^{n_0}}(\bar{T}^{(n_0)}) \cdot l_{\rho_{\mathcal{B}}}(J_{t+n_0}),\label{lip-module-Bellman}
	\end{align}
	where $l_{\rho_{\mathcal{B}}}^{{\mathcal{I}}^{\tau}}(\bar{T}^{(\tau)})$ is the supremum of the Lipschitz module $l_{\rho_{\mathcal{B}}}(\bar{T}^{(\tau)})$ over actions:
	\begin{align}
		l_{\rho_{\mathcal{B}}}^{{\mathcal{I}}^{\tau}}(\bar{T}^{(\tau)}) \coloneqq \sup_{i_{t:t+\tau-1}}\sup_{b_t\neq b'_t}\frac{K_{\rho_{\mathcal{B}}}(\bar{T}^{(\tau)}(db_{t+\tau}|b_t,i_{t:t+\tau-1}),\bar{T}^{(\tau)}(db_{t+\tau}|b'_t,i_{t:t+\tau-1}))}{\rho_{\mathcal{B}}(b_t,b'_t)}, \quad 0 \leq \tau \leq n_0.
	\end{align}
	Applying Lemma~\ref{lemma:lip_module_bound}, we deduce that
	for $n_i= i n_0$ with $0 \le i < k,$
	\begin{align}
		l_{\rho_{\mathcal{B}}}(J_{n_i})
		&\leq r_{\max} \cdot \sum_{\tau=0}^{n_0-1}\beta^\tau l_{\rho_{\mathcal{B}}}^{{\mathcal{I}}^{\tau}}(\bar{T}^{(\tau)})+\beta^{n_0} \cdot \gamma  \cdot l_{\rho_{\mathcal{B}}}(J_{n_{i+1}})\\
		&\leq r_{\max} \cdot \sum_{\tau=0}^{n_0-1}[C_3 \alpha^\tau + C_4]+\beta^{n_0} \cdot \gamma  \cdot l_{\rho_{\mathcal{B}}}(J_{n_{i+1}})\\
		&\leq \eta+\beta^{n_0}\gamma \cdot l_{\rho_{\mathcal{B}}}(J_{n_{i+1}}),
	\end{align}
	where
	\begin{align} \label{eq:eta}
		\eta = r_{\max} \cdot \left( \frac{C_3}{1-\alpha} + C_4 n _0\right),
	\end{align}
	and $C_3, C_4, n_0, \alpha$ are given in Lemma~\ref{lemma:lip_module_bound}.
	Iterating over $i$ and using $l_{\rho_{\mathcal{B}}}(J_{n}) = l_{\rho_{\mathcal{B}}}(J_{k n_0}) = r_{\max}$, we obtain
	\begin{align}
		l_{\rho_{\mathcal{B}}}(J_{0})\leq \eta \cdot \frac{1-\left(\beta^{n_0}\gamma \right)^{k}}{1-\beta^{n_0}\gamma} + \left(\beta^{n_0}\gamma \right)^{k} \cdot r_{\max}.
	\end{align}
	Recall that for $n=k n_0,$ $V_{n,\beta}^*(b) = V_{k n_0,\beta}^*(b) = J_{0}(b)$. Since $\beta <1$ and $\gamma<1$, we then get
	\begin{align}\label{lip-module-infinite}
		\liminf\limits_{k\rightarrow\infty}l_{\rho_{\mathcal{B}}}(V_{k n_0,\beta}^*) \leq \frac{\eta}{1-\gamma}.
	\end{align}
	Together with inequalities \eqref{inequ:span-bound} and \eqref{lip-module-average}, we obtain:
	\begin{align} \label{eq:bound-v-beta}
		|v_{\beta}(b)| \leq 2 l_{\rho_{\mathcal{B}}}(V_{\beta}^*) \leq 2\liminf\limits_{j\to\infty}l_{\rho_{\mathcal{B}}}(V_{j,\beta}^*) \leq \frac{2\eta}{1-\gamma} \quad \text{for any $b \in \mathcal{B},$}
	\end{align}
	where $\eta$ and $\gamma$ are independent of $\beta$. This proves the uniform boundedness of the bias functions $(v_{\beta})$. It then follows from Theorem 7 in \cite{hsu2006existence} that there exists a constant $\rho^*$ and a bounded continuous function $v$ satisfying the Bellman optimality equation \eqref{equ:Bellman-thm}, where
	\begin{align}
		\rho^* =\lim_{\beta \to 1-} (1-\beta)V_{\beta}^{*}(b),  \quad \text{and} \quad
		v(b)=\lim_{\beta \to 1-} v_{\beta}(b), \quad b\in\mathcal{B}. \label{eq:v-limit}
	\end{align}
	The proof is therefore complete. \Halmos

	\subsection{Proof of Proposition \ref{prop:span-uni-bound}}
	Recall from \eqref{eq:v-limit} that the bias function $v$ satisfies the relation
	$v(b)=\lim_{\beta \to 1-} v_{\beta}(b)$ for $b\in\mathcal{B}$.
	For two different belief states $b^1$ and $b^2$, we obtain from \eqref{def:discount bias func} that
	\begin{align} \label{eq:v-V}
		\left|v_\beta(b^1)-v_\beta(b^2)\right| =  \left|V^*_\beta(b^1)-V^*_\beta(b^2)\right| \le l_{\rho_{\mathcal{B}}}(V^*_{\beta}) \cdot ||b^1-b^2||_1.
	\end{align}
	Applying
	Lemma 3.2(a) \citep{hinderer2005lipschitz} and inequalities \eqref{eq:bound-v-beta} and \eqref{eq:v-V}, we have
	\begin{align}
		l_{\rho_{\mathcal{B}}}(v)\leq \lim_{\beta \to 1-}l_{\rho_{\mathcal{B}}}(v_{\beta})=\lim_{\beta \to 1-}l_{\rho_{\mathcal{B}}}(V_{\beta}^*) \leq \frac{\eta}{1-\gamma}.
	\end{align}
	Thus, from the definition of Lipschitz module \eqref{def:lip-module} and the definition $\text{span}(v) :=\max_{b \in \mathcal{B}}v(b)-\min_{b\in\mathcal{B}}v(b)$, we can deduce that for a POMDP model with $\epsilon=\min\limits_{i \in \mathcal{I}} \min\limits_{m,n \in \mathcal{M}}P^{(i)}(m,n)>0$, the bias span is bounded by
	\begin{align}
		\text{span}(v) \leq \frac{2 \eta}{1 - \gamma},
	\end{align}
	where $\eta$ is given in \eqref{eq:eta}, and $\gamma$ is given in Lemma~\ref{lemma:lip_module_bound}. Simplifying the expression, we obtain
	\begin{align} \label{eq:span-v-eps}
		\text{span}(v) \leq D(\epsilon):= \frac{8r_{\max}\left( \frac{2}{(1-\alpha(\epsilon))^2}+(1+\alpha(\epsilon)) \log_{\alpha(\epsilon)} \frac{1-\alpha(\epsilon)}{8}\right)}{1-\alpha(\epsilon)},
	\end{align}
	where with slight abuse of notations we use $\alpha(\epsilon)=\frac{1-2 \epsilon}{1 - \epsilon} \in (0,1)$ to emphasize its dependency on $\epsilon.$
	By Proposition~\ref{prop:spectral}, we can choose a sufficiently large $\tau_1$ so that the optimistic model with parameters $(\mathcal{P}_k,\Omega_k)$ in each episode $k \ge 1$ satisfies $\min\limits_{i \in \mathcal{I}} \min\limits_{m,n \in \mathcal{M}}P^{(i)}(m,n) \ge \epsilon/2>0$. Hence,
	we have for all $k,$
	\begin{align} \label{eq:lip-v-k}
		\text{span}(v_k) \leq D:=D(\epsilon/2).
	\end{align}
	The proof is therefore complete. \Halmos

	\subsection{Proof of Lemma~\ref{lemma:lip_module_bound}}\label{subsec: proof_lip_module_bound}
	
	To bound the Lipschitz module of the ($n-$step) belief transition kernel $\bar{T}^{(n)}$, we use the definition \eqref{def:lip-module-transi-kernel} and
	bound the following Kantorovich metric:
	\begin{align}
		&K\left(\bar{T}^{(n)}(db'|b^1,i_{0:n-1}),\bar{T}^{(n)}(db'|b^2,i_{0:n-1})\right)\\
		&=\sup_f\left\{\left|\int_{\mathcal{B}} f(b')\bar{T}^{(n)}(db'|b^1,i_{0:n-1})-\int_{\mathcal{B}} f(b')\bar{T}^{(n)}(db'|b^2,i_{0:n-1})\right|,f\in \text{Lip}_1\right\}\\
		&=\sup_f\left\{\left|\int_{\mathcal{B}} f(b')\bar{T}^{(n)}(db'|b^1,i_{0:n-1})-\int_{\mathcal{B}} f(b')\bar{T}^{(n)}(db'|b^2,i_{0:n-1})\right|,f\in \text{Lip}_1,||f||_\infty\leq 1\right\}.\\\label{equ:Kan_metric}
	\end{align}
	The last equality follows from the fact that for any $f\in \text{Lip}_1$, we have $|f(b)- f(b')| \le ||b-b'||_1 \le 2$ for $b, b' \in \B,$ and hence we can find a constant $c$ such that $||f + c||_\infty \leq 1$.
	
	To facilitate the analysis, we introduce a few definitions and notations. Define the $n$-step (conditional) observation kernel $\bar{Q}^{(n)}(\cdot |b,i_{0:n-1})$, which is a probability measure on $\mathcal{O}^n$:
	\begin{align}\label{def:n-belief-reward-kernel}
		\bar{Q}^{(n)}(\textbf{A}_1\times...\times \textbf{A}_n|b,i_{0:n-1}) \coloneqq \Prob\left((o_1,\dots,o_n)\in \textbf{A}_1\times...\times \textbf{A}_n|b,i_{0:n-1}\right), \quad \textbf{A}_i \subset \mathcal{O}.
	\end{align}
	Similar as the one-step belief updating function in \eqref{def:forward-kernel}, we also define the $n$-step forward kernel $H^{(n)}$ for the belief states so that $b_{n} = H^{(n)}(b,i_{0:n-1},o_{1:n}).$
	It is straightforward to check that for $\bar{T}^{(n)}$ defined in \eqref{def:n-trans-ker} we have
	\begin{align}
		\bar{T}^{(n)}(\textbf{A}|b,i_{0:n-1})=\int_{\mathcal{O}^n}\1_{\{H^{(n)}(b,i_{0:n-1},o_{1:n})\in \textbf{A}\}}\bar{Q}^{(n)}(\prod_{t=1}^{n}do_t|b,i_{0:n-1}), \quad \textbf{A} \subset \mathcal{B}.
	\end{align}
	
	Then to bound the Kantorovich metric in \eqref{equ:Kan_metric}, we can compute
	\begin{align}
		&\left|\int_{\mathcal{B}} f(b')\bar{T}^{(n)}(db'|b^1,i_{0:n-1})-\int_{\mathcal{B}} f(b')\bar{T}^{(n)}(db'|b^2,i_{0:n-1})\right|\\
		&=\left|\int_{\mathcal{O}^n} f(H^{(n)}(b^1,i_{0:n-1},o_{1:n}))\bar{Q}^{(n)}(\prod_{t=1}^{n}do_t|b^1,i_{0:n-1})-\int_{\mathcal{O}^n} f(H^{(n)}(b^2,i_{0:n-1},o_{1:n}))\bar{Q}^{(n)}(\prod_{t=1}^{n}do_t|b^2,i_{0:n-1})\right|\\
		&\leq\left|\int_{\mathcal{O}^n} f(H^{(n)}(b^1,i_{0:n-1},o_{1:n}))(\bar{Q}^{(n)}(\prod_{i=1}^{n}do_t|b^1,i_{0:n-1})-\bar{Q}^{(n)}(\prod_{t=1}^{n}do_t|b^2,i_{0:n-1},o_{1:n})) \right| \\
		&\quad \quad +\left|\int_{\mathcal{O}^n} (f(H^{(n)}(b^1,i_{0:n-1},o_{1:n}))- f(H^{(n)}(b^2,i_{0:n-1},o_{1:n})))\bar{Q}^{(n)}(\prod_{t=1}^{n}do_t|b^2,i_{0:n-1})\right|.\label{bound-Kan-metric}
	\end{align}
	
	We first bound the second term of \eqref{bound-Kan-metric}.
	By Lemma \ref{lemma:bt_exponential_decay}, for two different initial beliefs $b^1$ and $b^2$, we have
	\begin{equation}
		|H^{(n)}(b^1,i_{0:n-1},o_{1:n})-H^{(n)}(b^2,i_{0:n-1},o_{1:n})|\leq C_3\alpha^{n}||b^1-b^2||_1,
	\end{equation}
	where $C_3=\frac{2(1-\epsilon)}{\epsilon}, \alpha=1-\frac{\epsilon}{1-\epsilon}$, and $\epsilon=\min\limits_{i \in \mathcal{I}} \min\limits_{m,n \in \mathcal{M}}P^{(i)}(m,n)$. It follows that the second term of \eqref{bound-Kan-metric} can be bounded by
	\begin{align}
		&\left|\int_{\mathcal{O}^n} (f(H^{(n)}(b^1,i_{0:n-1},o_{1:n}))- f(H^{(n)}(b^2,i_{0:n-1},o_{1:n})))\bar{Q}^{(n)}(\prod_{t=1}^{n}do_t|b^2,i_{0:n-1})\right|\\
		&\leq \int_{\mathcal{O}^n} \left|H^{(n)}(b^1,i_{0:n-1},o_{1:n})- H^{(n)}(b^2,i_{0:n-1},o_{1:n})\right| \bar{Q}^{(n)}(\prod_{t=1}^{n}do_t|b^2,i_{0:n-1})\\
		&\leq C_3\alpha^{n}||b^1-b^2||_1,\label{bound-Kan-metric-p2}
	\end{align}
	where we use the fact that $f\in \text{Lip}_1$.
	
	We next bound the first term of \eqref{bound-Kan-metric}.
	From the fact that the observations are finite, we know that the $n$-step observation kernel
	$\bar{Q}^{(n)}(\prod_{t=1}^{n}do_t|b_0,i_{0:n-1})$ is a measure with probability mass function denoted by $\bar{Q}^{(n)}( o_{1:n}|b_0,i_{0:n-1})$ (with slight abuse of notations), and
	\begin{align}
		\bar{Q}^{(n)}( o_{1:n}|b_0,i_{0:n-1})&=\sum_{m\in\mathcal{M}}\Prob(M_0=m)\Prob(o_{1:n}|M_0=m,i_{0:n-1}) \\
		&=\sum_{m\in\mathcal{M}}b_0(m)\Prob(o_{1:n}|M_0=m,i_{0:n-1}).
	\end{align}
	Then we have
	\begin{align}
		&\left|\int_{\mathcal{O}^n} f(H^{(n)}(b^1,i_{0:n-1},o_{1:n}))\left(\bar{Q}^{(n)}(\prod_{t=1}^{n}do_t|b^1,i_{0:n-1})-\bar{Q}^{(n)}(\prod_{t=1}^{n}do_t|b^2,i_{0:n-1})\right)\right|\\
		&=\left|\sum_{m=1}^{M}(b^1(m)-b^2(m))\sum_{ o_{1:n} \in \mathcal{O}^n}f(H^{(n)}(b^1,i_{0:n-1},o_{1:n}))\Prob(o_{1:n} |M_0=m,i_{0:n-1})\right|\\
		&=\left|\sum_{m=1}^{M}\left(b^1(m)-b^2(m)\right)g(m)\right|\\
		&=\left|\sum_{m=1}^{M}\left(b^1(m)-b^2(m)\right)\left(g(m)-\frac{\max_m g(m)+\min_m g(m)}{2}\right)\right|\\
		&\leq \left \|b^1-b^2\right \|_1 \cdot \left \|g(m)-\frac{\max_m g(m)+\min_m g(m)}{2}\right \|_\infty\\
		&= \left \|b^1-b^2\right \|_1 \cdot \frac{1}{2}\left(\max_m g(m)-\min_m g(m)\right),\label{bound-Kan-metric-p1-0}
	\end{align}
	where we have used the fact that $\sum_{m=1}^{M}\left(b^1(m)-b^2(m)\right)=0$, and the function
	$g$ is defined as follows:
	\begin{equation}
		g(m) \coloneqq \sum_{ o_{1:n} \in \mathcal{O}^n}f(H^{(n)}(b^1,i_{0:n-1},o_{1:n}))\Prob(o_{1:n} |M_0=m,i_{0:n-1}).
	\end{equation}  
	From the equation above, it is clear that the quantity $\frac{1}{2}\left(\max_m g(m)-\min_m g(m)\right) \le 1$, because $||f ||_\infty \leq 1$. However to prove Lemma~\ref{lemma:lip_module_bound}, we need a sharper bound so that we can find a constant $C_4<1$ (that is independent of $b^1, n$ and $i_{0:n-1}$) with
	\begin{equation}\label{eq:C4-bound}
		\frac{1}{2}\left(\max_m g(m)-\min_m g(m)\right) \le C_4 <1.
	\end{equation}
	
	Suppose \eqref{eq:C4-bound} holds. Then
	on combining \eqref{bound-Kan-metric}, \eqref{bound-Kan-metric-p2} and \eqref{bound-Kan-metric-p1-0}, we obtain
	\begin{align}\label{inequ:Kan_insider}
		\left|\int_{\mathcal{B}} f(b')\bar{T}^{(n)}(db'|b^1,i_{0:n-1})-\int_{\mathcal{B}} f(b')\bar{T}^{(n)}(db'|b^2,i_{0:n-1})\right|
		\leq C_3\alpha^{n}||b^1-b^2||_1+C_4||b^1-b^2||_1,
	\end{align}
	where $\alpha<1,C_4<1$. It then follows
	from equation \eqref{equ:Kan_metric} that the Kantorovich metric is bounded by
	\begin{align}\label{inequ:Kantorovich metric}
		K\left(\bar{T}^{(n)}(db'|b^1,i_{0:n-1}),\bar{T}^{(n)}(db'|b^2,i_{0:n-1})\right) \leq C_3\alpha^{n}||b^1-b^2||_1+C_4||b^1-b^2||_1, \text{ for any }i_{0:n-1},
	\end{align}
	where $C_3=\frac{2(1-\epsilon)}{\epsilon}, \alpha=1-\frac{\epsilon}{1-\epsilon}$, and $\epsilon=\min\limits_{i \in \mathcal{I}} \min\limits_{m,n \in \mathcal{M}}P^{(i)}(m,n)>0$.
	So $\bar{T}^{(n)}$ is Lipschitz uniformly in any $i_{0:n-1}$, and
	its Lipschitz module can be bounded as follows:
	\begin{align}
		l^{\mathcal I^{n}}_{\rho_{\mathcal{B}}}(\bar{T}^{(n)}):=\sup_{i_{0:n-1}}\sup_{b^1\neq b^2}\frac{K\left(\bar{T}^{(n)}(db'|b^1,i_{0:n-1}),\bar{T}^{(n)}(db'|b^2,i_{0:n-1})\right)}{\rho_{\mathcal{B}}(b^1,b^2)} \le C_3\alpha^{n}+C_4.
	\end{align}
	If we choose $n=n_0:=\lceil \log_{\alpha}\frac{1-C_4}{2C_3} \rceil$, so that
	$C_3\alpha^{n_0}+C_4<\frac{1}{2}(1+C_4)\coloneqq\gamma<1$, then
	we obtain the desired result $
	l^{\mathcal I^{n_0}}_{\rho_{\mathcal{B}}}(\bar{T}^{(n_0)})< \gamma.$
	
	It remains to prove \eqref{eq:C4-bound}.
	Since the set $\mathcal{M}=\{1, \ldots, M\}$ is finite, we pick $m^* \in \argmin \limits_{m \in \mathcal{M}} g(m), \hat m \in \argmax \limits_{m \in \mathcal{M}} g(m)$. We have
	\begin{align} \label{eq:diff-g}
		&\frac{1}{2}\left(\max_m g(m)-\min_m g(m)\right)\\
		&= \frac{1}{2}\sum_{o_{1:n} \in \mathcal{O}^n}f(H^{(n)}(b^1,i_{0:n-1},o_{1:n}))\left(\Prob(o_{1:n}|M_0=\hat m,i_{0:n-1})-\Prob(o_{1:n}|M_0=m^*,i_{0:n-1})\right)\\
		&\leq \frac{1}{2} \sum_{o_{1:n} \in \mathcal{O}^n } \left|\Prob(o_{1:n}|M_0=\hat m,i_{0:n-1})-\Prob(o_{1:n}|M_0=m^*,i_{0:n-1})\right|,
	\end{align}
	where the inequality follows from H\" older's inequality with $||f||_\infty\leq 1$.
	We can compute
	\begin{align}
		&\Prob(o_{1:n}|M_0=m_0,i_{0:n-1})\\
		&=\sum_{m_{1:n}\in\mathcal{M}^n}\Prob(o_{1:n}|M_0=m_0,i_{0:n-1},M_{1:n}=m_{1:n})\cdot \Prob(M_{1:n}=m_{1:n}|M_0=m_0,i_{0:n-1})\\
		&=\sum_{m_{1:n}\in\mathcal{M}^n}\Prob(o_{1:n}|i_{0:n-1},m_{1:n})\cdot \Prob(m_{1:n}|M_0=m_0,i_{0:n-1})\\
		&=\sum_{m_{1:n}\in\mathcal{M}^n}\left(\prod_{t=1}^{n}\Prob(o_t|m_t,i_{t-1})\right) \cdot \left(\prod_{t=0}^{n-1}\Prob(m_{t+1}|m_t,i_{t})\right),
	\end{align}
	where the second equality follows from the fact that given sequences $\{i_{0:n-1}\}$ and $\{m_{1:n}\}$, $\{o_{1:t}\}$ is independent from $M_0$, and the last equality holds due to the conditional independence. Using the assumption that $\epsilon=\min\limits_{i \in \mathcal{I}} \min\limits_{m,n \in \mathcal{M}}P^{(i)}(m,n)>0$, we can then infer that for any $\{o_{1:n}\},  \{i_{0:n-1}\}$,
	\begin{align}
		&\Prob(o_{1:n}|M_0=m^*,i_{0:n-1})\\
		&=\sum_{m_{1:n}\in\mathcal{M}^n}\left(\prod_{t=1}^{n}\Prob(o_t|m_t,i_{t-1})\right) \cdot \left(\prod_{t=1}^{n-1}\Prob(m_{t+1}|m_t,i_{t})\right) \cdot P^{(i_0)}(m^*,m_1) \\
		& \ge \sum_{m_{1:n}\in\mathcal{M}^n}\left(\prod_{t=1}^{n}\Prob(o_t|m_t,i_{t-1})\right) \cdot \left(\prod_{t=1}^{n-1}\Prob(m_{t+1}|m_t,i_{t})\right) \cdot P^{(i_0)}( \hat m,m_1) \cdot \frac{\epsilon}{1 - \epsilon} \\
		& = \Prob(o_{1:n}|M_0= \hat m,i_{0:n-1}) \cdot \frac{\epsilon}{1 - \epsilon}.
	\end{align}
	It follows that
	\begin{align*}
		& \left|\Prob(o_{1:n}|M_0=\hat m,i_{0:n-1})-\Prob(o_{1:n}|M_0=m^*,i_{0:n-1})\right| \\
		& \le \max\left\{\left(1- \frac{\epsilon}{1 - \epsilon} \right) \Prob(o_{1:n}|M_0=\hat m,i_{0:n-1}), \Prob(o_{1:n}|M_0=m^*,i_{0:n-1}) \right\} \\
		& \le \left(1- \frac{\epsilon}{1 - \epsilon} \right) \Prob(o_{1:n}|M_0=\hat m,i_{0:n-1}) + \Prob(o_{1:n}|M_0=m^*,i_{0:n-1}).
	\end{align*}
	Then we can obtain from \eqref{eq:diff-g} that
	\begin{align}
		&\frac{1}{2}\left(\max_m g(m)-\min_m g(m)\right)\\
		& \leq \frac{1}{2} \sum_{o_{1:n} \in \mathcal{O}^n} \left[
		\left(1- \frac{\epsilon}{1 - \epsilon} \right) \Prob(o_{1:n}|M_0=\hat m,i_{0:n-1}) + \Prob(o_{1:n}|M_0=m^*,i_{0:n-1}) \right] \\
		& = \alpha/2 + 1/2 := C_4 <1,
	\end{align}
	where $\alpha= 1 - \frac{\epsilon}{1 - \epsilon} \in (0,1).$ The proof is complete. \Halmos

	\section{Proof of Lemma \ref{lemma:view_to_f_P}}
	Firstly, consider the $(o,m)$ entry of matrix $A_2^{(i)}$, we have
	\begin{align}
		A_2^{(i)}(o,m)&=\Prob(v^{(i)}_{2,t}=\bm{e}_o|m_{t}=m,i_{t}=i)\\
		&=\Prob(o_{t}=o|m_{t}=m,i_{t}=i)\\
		&=\Omega(o|m,i),
	\end{align}
	where the last equality is due to the deterministic policy $i_t=i_{t-1}=i$.
	
	Secondly, for the $(o,m)$ entry of matrix $A_3^{(i)}$, we can get:
	\begin{align}
		A_3^{(i)}(o,m)&=\Prob(v^{(i)}_{3,t}=\bm{e}_o|m_t=m,i_t=i)\\
		&= \Prob(o_{t+1}=o|m_t=m,i_t=i)\\
		&=\sum_{m' \in \mathcal{M}}\Prob(o_{t+1}=o|m_t=m,i_t=i,m_{t+1}=m')\Prob(m_{t+1}=m'|m_t=m,i_t=i)\\
		&=\sum_{m' \in \mathcal{M}}\Prob(o_{t+1}=o|i_t=i,m_{t+1}=m')\Prob(m_{t+1}=m'|m_t=m,i_t=i),
	\end{align}
	where the last equality follows from the fact that given $m_{t+1}$ and $i_{t}$, $o_{t+1}$ and $m_t$ are independent.
	Then we have
	\begin{align}
		A_3^{(i)}(o,m)=\sum_{m' \in \mathcal{M}}A_2^{(i)}(o,m')P^{(i)}(m,m')=\left[A_2^{(i)}\left(P^{(i)}\right)^{\top}\right](o,m).
	\end{align}
	Thus, the transition matrix can be recovered by $P^{(i)}=\left(\left(A_2^{(i)}\right)^\dagger A_3^{(i)}\right)^{\top}$. \Halmos
	
	\section{Proof of Proposition \ref{prop:spectral}} \label{sec:proof-prop-spectral}
	Recall three matrices defined in \eqref{multi-view}:
	\begin{align}
		&A_1^{(i)}(o,m)=\Prob(v^{(i)}_{1,t}=\bm{e}_o|m_t=m,i_t=i),\\
		&A_2^{(i)}(o,m)=\Prob(v^{(i)}_{2,t}=\bm{e}_o|m_t=m,i_t=i),\\
		&A_3^{(i)}(o,m)=\Prob(v^{(i)}_{3,t}=\bm{e}_o|m_t=m,i_t=i).
	\end{align}
	Following from Theorem 3, Lemma 5 and Lemma 8 in \citet{azizzadenesheli2016reinforcement}, we know that for fixed $\delta \in (0,1)$, for any action $i \in \mathcal{I}$, when the number of samples $N^{(i)} \ge N_0^{(i)}$, where
	\begin{align}
		N_0^{(i)}:= \max \left\{\frac{4}{\left(\sigma_{3,1}^{(i)}\right)^2}, \left(\frac{G^{(i)}\frac{2\sqrt{2}+1}{1-\theta^{(i)}}}{\left(\omega^{(i)}\right)_{\min}\left(\sigma^{(i)}\right)^2}\right)^2 \max\left\{\frac{16 \times M^{1/3}}{C_0^{2/3}\left(\left(\omega^{(i)}\right)_{\min}\right)^{1/3}},\frac{2\sqrt{2}M}{C_0^2\left(\omega^{(i)}\right)_{\min}\left(\sigma^{(i)}\right)^2},4\right\}\right\}\log \left(\frac{6(O^2+O)}{\delta}\right),
	\end{align}
	then with probability $1-\delta$, the spectral estimators $\hat{\Omega}, \hat{P}$ have the following guarantee:
	\begin{align}
		||\Omega(\cdot|m,i)-\hat{\Omega}(\cdot|m,i)||_1 &\leq C_1^{(i)} \sqrt{\frac{\log\left(\frac{6(O^2+O)}{\delta}\right)}{N^{(i)}}},\\
		||P^{(i)}(m,:)- \hat{P}^{(i)}(m,:)||_2&\leq C_2^{(i)}\sqrt{\frac{\log\left(\frac{6(O^2+O)}{\delta}\right)}{N^{(i)}}},
	\end{align}
	for $i \in \mathcal{I}$ and $m \in \mathcal{M}$ up  to permutation, with
	\begin{align}\label{eq:defC}
		C_1^{(i)}&=\frac{21\sqrt{O}}{\sigma_{3,1}^{(i)}}C_{12}^{(i)},\\
		C_2^{(i)}&=\frac{4}{\sigma_{\min}\left(A_2^{(i)} \right)}\left(\sqrt{M} +\frac{21M}{\sigma_{3,1}^{(i)}}\right)C_{12}^{(i)},\\
		C_{12}^{(i)}&= 2G^{(i)}\frac{2\sqrt{2}+1}{\left(1-\theta^{(i)})\right)\left(\left(\omega^{(i)}\right)_{\min}\right)^{1/2}}\left(1+\frac{8\sqrt{2}}{\left(\left(\omega^{(i)}\right)_{\min}\right)^{2}\left(\sigma^{(i)}\right)^3}+\frac{256}{\left(\left(\omega^{(i)}\right)_{\min}\right)^{2}\left(\sigma^{(i)}\right)^2}\right).
	\end{align}
	Here, $C_0$ is a numerical constant,
	$\sigma_{3,1}^{(i)}$ is the smallest nonzero singular value of the covariance matrix $W_{3,1}^{(i)}$, and $\sigma^{(i)}=\min\{\sigma_{\min}(A_1^{(i)}),\sigma_{\min}(A_2^{(i)}),\sigma_{\min}(A_3^{(i)})\}$ with $\sigma_{\min}(A_j^{(i)})$ represents the smallest nonzero singular value of the matrix $A_j^{(i)}$, for $j=1,2,3$. Moreover, $\left(\omega^{(i)}\right)_{\min}=\min \limits_m \omega^{(i)}(m)\ge \epsilon$,
	where $\omega^{(i)}$ is the stationary distribution of the
	geometrically ergodic Markov chain with transition matrix $P^{(i)}$. Furthermore, $G^{(i)}$ and $\theta^{(i)}$ are the mixing rate parameters of the Markov chain such that
	\begin{align}
		\sup_{m_0}||m_0 [P^{(i)}]^t -\omega^{(i)}||_{TV}\leq G^{(i)}\left(\theta^{(i)})\right)^{t},
	\end{align}
	where $TV$ stands for the total variation distance.
	Using Assumption~\ref{assum:trans_matrix_min},
	one can take $G^{(i)} = 2$ and have the (crude) bound $\theta^{(i)} \le 1 - \epsilon$ with $\epsilon \in (0,1)$, see e.g. Theorems 2.7.2 and 2.7.4 in \cite{krishnamurthy2016partially}.
	Finally, setting $C_1\coloneqq\max \limits_{i \in \mathcal{I}} C_1^{(i)}$ and $C_2\coloneqq\max\limits_{i \in \mathcal{I}} C_2^{(i)}$, we complete the proof. \Halmos

	\section{Proof of Proposition \ref{prop:lip_bt}}\label{sec:proof_lip_bt}
	We use a similar approach as the proof of Proposition 3 in \citet{de2017consistent}, where they bound the filtering error in hidden Markov models. We extend it to the setting of POMDP.
	
	We first introduce some notations to facilitate the presentation. Denote by $H_t$ the $t-$th iteration of the Bayesian filtering recursion \eqref{def:forward-kernel} under the true model parameters $P^{(i)}$ and $\Omega$, so that we have ${b}_{t+1}={H}_{t+1} {b}_{t}$, where
	\begin{align}
		b_{t+1}(m_{t+1})&=\frac{\Omega(o_{t+1}|m_{t+1},i_t)\sum \limits_{m_t\in\mathcal{M}}P^{(i_t)}(m_t,m_{t+1})b_t(m_t)}{\sum \limits_{m_{t+1}\in\mathcal{M}}\Omega(o_{t+1}|m_{t+1},i_t)\sum\limits_{m_t\in\mathcal{M}}P^{(i_t)}(m_t,m_{t+1})b_t(m_t)},
	\end{align}
	and we omit the dependency of $H_t$ on the action $i_t$ and the observation $o_{t+1}$ to simplify notations.
	Similarly, denote by $\hat H_t$ the approximation of $H_t$ obtained by replacing $P^{(i)}$ and $\Omega$ by the estimators $\hat P^{(i)}$ and $\hat \Omega$, so that $\hat {b}_{t+1} =\hat {H}_{t+1} \hat {b}_{t}$, where
	\begin{align}
		\hat b_{t+1}(m_{t+1})&=\frac{\hat \Omega(o_{t+1}|m_{t+1},i_t)\sum \limits_{m_t\in\mathcal{M}} \hat P^{(i_t)}(m_t,m_{t+1})\hat b_t(m_t)}{\sum \limits_{m_{t+1}\in\mathcal{M}} \hat \Omega(o_{t+1}|m_{t+1},i_t)\sum\limits_{m_t\in\mathcal{M}}\hat P^{(i_t)}(m_t,m_{t+1}) \hat b_t(m_t)}.
	\end{align}

	Using the fact that $b_0 = \hat b_0$, we can compute for $t \geq 1$,
	\begin{align}
		&b_t-\hat{b}_t\\
		&=H_t b_{t-1}-\hat{H}_t \hat{b}_{t-1}\\
		&=H_t b_{t-1}-H_t \hat{b}_{t-1}+H_t \hat{b}_{t-1}-\hat{H}_t \hat{b}_{t-1}\\
		&=\sum_{l=1}^{t-1}[H_t H_{t-1}\dots H_l \hat{b}_{l-1}-H_t H_{t-1} \dots H_{l+1} \hat{b}_{l}]+[H_t \hat{b}_{t-1}-\hat{H}_t \hat{b}_{t-1}]\\
		& =\sum_{l=1}^{t-1}[H_t H_{t-1}\dots H_l \hat{b}_{l-1}-H_t H_{t-1} \dots H_{l+1} \hat{H}_l \hat{b}_{l-1}]+[H_t \hat{b}_{t-1}-\hat{H}_t \hat{b}_{t-1}],\label{error:filtering-distr}
	\end{align}
	where the third equality is due to the telescoping sum.
	
	We first bound the second term of equation \eqref{error:filtering-distr}. \rev{Note we have the simple inequality $\left|\frac{A}{B} - \frac{C}{D} \right| \le \left|\frac{A}{B} - \frac{C}{B}\right| + \left| \frac{C}{B} - \frac{C}{D} \right|
		= \left| \frac{A- C}{B} \right| + \left| \frac{C}{B}\right| \times \left| \frac{B-D}{D} \right |$ for real numbers $A, B, C, D$ with $B, D \ne 0.$ From the definitions of $H_t$ and $\hat{H}_t$, we can then use this inequality to verify that
		for any bounded function $h$ on $\mathcal{M},$ we have $|\langle H_t \hat{b}_{t-1}, h \rangle - \langle \hat{H}_t \hat{b}_{t-1}, h \rangle | \le S_1 + S_2$,
		where the notation $\langle  , \rangle$ denotes the inner product, and }
	\begin{align*}
		S_1 &:= \left|\frac{\sum\limits_{m_{t-1},m_t} g(m_{t-1},m_t,i_{t-1},o_t) \cdot
			\hat{b}_{t-1}(m_{t-1})h(m_t)}{\sum\limits_{m_{t-1},m_t}\Omega(o_t|m_t,i_{t-1})P^{(i_{t-1})}(m_{t-1},m_t)\hat{b}_{t-1}(m_{t-1})}\right|, \\
		S_2 &:= \left|\frac{\sum \limits_{m_{t-1}, m_t} \hat{\Omega}(o_t|m_t,i_{t-1})\hat{P}^{(i_{t-1})}(m_{t-1},m_t)\hat{b}_{t-1}(m_{t-1})h(m_t)}{\sum\limits_{m_{t-1},m_t}\Omega(o_t|m_t,i_{t-1})P^{(i_{t-1})}(m_{t-1},m_t)\hat{b}_{t-1}(m_{t-1})} \right| \\
		& \quad \quad \times \left| \frac{\sum\limits_{m_{t-1},m_{t}} g(m_{t-1},m_t,i_{t-1},o_t) \cdot \hat{b}_{t-1}(m_{t-1})}{\sum \limits_{m_{t-1},m_t}\hat{\Omega}(o_t|m_t,i_{t-1})\hat{P}^{(i_{t-1})}(m_{t-1},m_t)\hat{b}_{t-1}(m_{t-1})}\right|,
	\end{align*}
	where $g(m_{t-1},m_t,i_{t-1},o_t) =\Omega(o_t|m_t,i_{t-1})P^{(i_{t-1})}(m_{t-1},m_t)- \hat{\Omega}(o_t|m_t,i_{t-1})\hat{P}^{(i_{t-1})}(m_{t-1},m_t)$.
	The term $S_1$ can be bounded as follows:
	\begin{align}
		S_1 &\leq \frac{\sum\limits_{m_{t-1},m_t} \left|\Omega(o_t|m_t,i_{t-1})- \hat{\Omega}(o_t|m_t,i_{t-1})\right|P^{(i_{t-1})}(m_{t-1},m_t)\hat{b}_{t-1}(m_{t-1})|h(m_t)|}{\sum\limits_{m_{t-1},m_t}\Omega(o_t|m_t,i_{t-1})P^{(i_{t-1})}(m_{t-1},m_t)\hat{b}_{t-1}(m_{t-1})}\\
		&\quad \quad + \frac{\sum\limits_{m_{t-1},m_t} \left|P^{(i_{t-1})}(m_{t-1},m_t)-\hat{P}^{(i_{t-1})}(m_{t-1},m_t) \right|\hat{\Omega}(o_t|m_t,i_{t-1})\hat{b}_{t-1}(m_{t-1})|h(m_t)|}{\sum\limits_{m_{t-1},m_t}\Omega(o_t|m_t,i_{t-1})P^{(i_{t-1})}(m_{t-1},m_t)\hat{b}_{t-1}(m_{t-1})}\\
		&\leq \frac{\max\limits_{m_t} \left \|\Omega(\cdot|m_t,i_{t-1})- \hat{\Omega}(\cdot |m_t,i_{t-1}) \right \|_1}{\xi}\cdot \|h\|_{\infty}+\frac{\max\limits_{m_{t-1}} \left\| P^{(i_{t-1})}(m_{t-1},:)-\hat{P}^{(i_{t-1})}(m_{t-1},:) \right\|_2}{\epsilon}\cdot \|h\|_{\infty},
	\end{align}
	where the last step is due to $\xi=\min\limits_{o \in \mathcal{O}}\min\limits_{i \in \mathcal{I}}\min\limits_{m \in \mathcal{M}} \sum\limits_{n \in \mathcal{M}}\Omega(o|n,i) P^{(i)}(m,n)>0$,
	$\epsilon=\min\limits_{i \in \mathcal{I}} \min\limits_{m,n \in \mathcal{M}} P^{(i)}(m,n)>0$, $\left|\Omega(o_t|m_t,i_{t-1})- \hat{\Omega}(o_t|m_t,i_{t-1})\right| \leq \max_{m_t} \left \|\Omega(\cdot|m_t,i_{t-1})- \hat{\Omega}(\cdot |m_t,i_{t-1}) \right \|_1$, and $\left|P^{(i_{t-1})}(m_{t-1},m_t)-\hat{P}^{(i_{t-1})}(m_{t-1},m_t) \right| \leq \max_{m_{t-1}} \left\| P^{(i_{t-1})}(m_{t-1},:)-\hat{P}^{(i_{t-1})}(m_{t-1},:) \right\|_2.$ The same upper bound holds for $S_2$. Thus, applying Lemma \ref{lemma:measure-holder}, we obtain
	\begin{align}
		&||H_t \hat{b}_{t-1}-\hat{H}_t \hat{b}_{t-1}||_1\\
		&	\quad \leq  2\frac{\max\limits_{m_t} \left \|\Omega(\cdot|m_t,i_{t-1})- \hat{\Omega}(\cdot |m_t,i_{t-1}) \right \|_1}{\xi}+2\frac{\max\limits_{m_{t-1}} \left\| P^{(i_{t-1})}(m_{t-1},:)-\hat{P}^{(i_{t-1})}(m_{t-1},:) \right\|_2}{\epsilon}.
	\end{align}

	We next bound the first term of \eqref{error:filtering-distr}. Note that $H_t H_{t-1}\dots H_l \hat{b}_{l-1}-H_t H_{t-1} \dots H_{l+1} \hat{H}_l \hat{b}_{l-1}$ can be viewed as the error between the corresponding belief states at time $t$ given two different belief states $H_l \hat{b}_{l-1}$ and $\hat{H}_l \hat{b}_{l-1}$ at time $l$ under the true model. Thus, we can infer from Lemma \ref{lemma:bt_exponential_decay} that
	\begin{align}
		\left\| H_t H_{t-1}\dots H_l \hat{b}_{l-1}-H_t H_{t-1} \dots H_{l+1} \hat{H}_l \hat{b}_{l-1} \right \|_1
		\leq C_3 \rho^{t-l}\left\|H_l \hat{b}_{l-1}-\hat{H}_l \hat{b}_{l-1}\right\|_1,
	\end{align}
	where $C_3=\frac{2(1-\epsilon)}{\epsilon} \ge 1$, and $\rho=1-\frac{\epsilon}{1-\epsilon} \in (0,1)$.
	So the first term of \eqref{error:filtering-distr} can be bounded by:
	\begin{align}
		\sum_{l=1}^{t-1}\left \|H_t H_{t-1}\dots H_l \hat{b}_{l-1}-H_t H_{t-1} \dots H_{l+1} \hat{H}_l \hat{b}_{l-1}\right \|_1
		\leq \sum_{l=1}^{t-1}C_3 \rho^{t-l}\left\|H_l \hat{b}_{l-1}-\hat{H}_l \hat{b}_{l-1}\right\|_1.
	\end{align}
	
	Thus, we have
	\begin{align}
		&\left \|b_t-\hat{b}_t\right \|_{1}\\
		&\leq \sum_{l=1}^{t-1}\left \|H_t H_{t-1}\dots H_l \hat{b}_{l-1}-H_t H_{t-1} \dots H_{l+1} \hat{H}_l \hat{b}_{l-1}\right \|_1+\left\|H_t \hat{b}_{t-1}-\hat{H}_t \hat{b}_{t-1}\right\|_{1}\\
		&	\leq  \sum_{l=1}^{t-1} C_3 \rho^{t-l}\left\|H_l \hat{b}_{l-1}-\hat{H}_l \hat{b}_{l-1}\right\|_1+\left\|H_t \hat{b}_{t-1}-\hat{H}_t \hat{b}_{t-1}\right\|_{1}\\
		&	\leq  \sum_{l=1}^{t}C_3 \rho^{t-l}\left\|H_l \hat{b}_{l-1}-\hat{H}_l \hat{b}_{l-1}\right\|_{1}\\
		&	\leq \sum_{l=1}^{t}2C_3 \rho^{t-l} \times \\
		& \left( \frac{\max_{m_l} \left \|\Omega(\cdot|m_l,i_{l-1})- \hat{\Omega}(\cdot |m_l,i_{l-1}) \right \|_1}{\xi}  +\frac{\max_{m_{l-1}} \left\| P^{(i_{l-1})}(m_{l-1},:)-\hat{P}^{(i_{l-1})}(m_{l-1},:) \right\|_2}{\epsilon}\right).
	\end{align}
	For any give sequence $\{i_{0:t-1},o_{1:t}\}_{t\geq 1}$, we have
	\begin{align}
		&\max_{m_l} \left \|\Omega(\cdot|m_l,i_{l-1})- \hat{\Omega}(\cdot |m_l,i_{l-1}) \right \|_1 \leq \max_{m \in \mathcal{M}, i\in \mathcal{I}}\left\|\Omega(\cdot|m,i)-\hat{\Omega}(\cdot|m,i)\right\|_1, \quad 1 \leq l \leq t,\\
		&\max_{m_{l-1}} \left\| P^{(i_{l-1})}(m_{l-1},:)-\hat{P}^{(i_{l-1})}(m_{l-1},:) \right\|_2 \leq \max_{m \in \mathcal{M}, i\in \mathcal{I}}\left\| P^{(i)}(m,:)-\hat{P}^{(i)}(m,:) \right\|_2, \quad 1 \leq l \leq t.
	\end{align}
	Therefore, we can obtain:
	\begin{align}
		||b_t-\hat{b}_t||_{1}
		\leq4\left( \frac{1-\epsilon}{\epsilon}\right)^2\left(\frac{\max\limits_{m \in \mathcal{M}, i\in \mathcal{I}}\left\|\Omega(\cdot|m,i)-\hat{\Omega}(\cdot|m,i)\right\|_1}{\xi}+\frac{\max\limits_{m \in \mathcal{M}, i\in \mathcal{I}}\left\| P^{(i)}(m,:)-\hat{P}^{(i)}(m,:) \right\|_2}{\epsilon}\right).
	\end{align}
	Letting $L_1=\frac{4(1-\epsilon)^2}{\epsilon^2 \xi}$ and
	$L_2=\frac{4(1-\epsilon)^2}{\epsilon^3}$, we have shown
	\begin{align}
		\Vert b_t -\hat{b}_t \Vert_1\leq L_1 \max\limits_{m \in \mathcal{M}, i\in \mathcal{I}}\left\|\Omega(\cdot|m,i)-\hat{\Omega}(\cdot|m,i)\right\|_1+L_2 \max\limits_{m \in \mathcal{M}, i\in \mathcal{I}} \left\| P^{(i)}(m,:)-\hat{P}^{(i)}(m,:) \right\|_2.
	\end{align}
	The proof is therefore complete. \Halmos
	
	\section{Proof of Theorem \ref{thm:upper_bound}} \label{sec:proof-complete}
	
	We use an approach inspired by \cite{jaksch2010near}.
	Fixing an arbitrary initial belief state $b$, recall the regret of a learning policy defined in \eqref{def:reg}.
	For our learning policy $\pi$ (Algorithm~\ref{alg:SEEU}), we proceed to bound
	\begin{align}
		&(T+1)\rho^* - \sum_{t=0}^T R_t^{\pi}(b)\\
		&=\left[(T+1)\rho^*-\sum_{t=0}^T\mathbb{E} \left[ R_t^{\pi}(b) |\mathcal{H}_t\right]\right]+\left[\sum_{t=0}^T\mathbb{E} \left[ R_t^{\pi}(b) |\mathcal{H}_t\right]-\sum_{t=0}^T R_t^{\pi}(b) \right],\label{reg}
	\end{align}
	where $\mathcal{H}_t=\{\pi_0,O^{\pi}_1,...,\pi_{t-1},O^{\pi}_{t}\} $ denotes the observable history.

	We first bound the second term of \eqref{reg}. Define a stochastic process $\{X_t\}_{t=0}^T$ as follows:
	\begin{align}
		&X_0=0,\\
		&X_t=\sum_{l=0}^{t}\left(\mathbb{E} \left[ R_l^{\pi}(b) |\mathcal{H}_l\right]-R_l^{\pi}(b)\right),\quad t \geq 1.
	\end{align}
	It is easy to check that $\{X_t : t \ge 0 \}$ is a martingale.
	Moreover, since rewards are non-negative and upper bounded by $r_{\max}$, we can use the simple inequality $|a-c| \le \max\{a, c\}$ for $a, c \ge 0$ and obtain
	$ |X_{t+1}-X_t|=|\mathbb{E} \left[ R_t^{\pi}(b) \left|\mathcal{H}_t\right]-R_t^{\pi}(b)\right|\leq r_{\max}.$
	Then we can apply the Azuma-Hoeffding inequality \citep{azuma1967weighted} to the martingale $\{X_t : t \ge 0 \}$ and obtain for $\delta \in (0,1),$
	\begin{align}\label{reg:expreward-reward}
		\Prob\left(\sum_{l=0}^{T}\left(\mathbb{E} \left[ R_l^{\pi}(b) |\mathcal{H}_l\right]-R_l^{\pi}(b)\right) \geq \sqrt{2Tr_{\max}\log\left(\frac{1}{\delta}\right)}\right) \leq \delta.
	\end{align}
	
	Next we bound the first term in Equation \eqref{reg}.
	Using the definition of $R_t^{\pi}(b)$ in \eqref{def:policy_reward}, then we can rewrite the first term in \eqref{reg}:
	\begin{align}\label{reg1}
		\sum_{t=0}^{T}\left(\rho^*-\mathbb{E} \left[ R_t^{\pi}(b) |\mathcal{H}_t\right]\right)=\sum_{t=0}^{T}(\rho^*-\E[R(M_t, \pi_t)|\mathcal{H}_{t}])=\sum_{t=0}^{T}(\rho^*-\bar{R}(b_t,\pi_t)),
	\end{align}
	where $\bar{R}(b, i)$ is defined in
	\eqref{def:belief-reward-func}.
	To bound \eqref{reg1}, we study the regret in exploration phases and exploitation phases separately. To this end,
	let $K$ be the total number of episodes from time $0$ to $T$. For each episode $k=1, 2, \cdots, K$, let $A_k, E_k$ be the exploration and exploitation phases, respectively.
	Then we have the following decomposition:
	\begin{align}\label{reg2}
		\sum_{t=0}^{T}\left(\rho^*-\mathbb{E} \left[ R_t^{\pi}(b) |\mathcal{H}_t\right]\right)=    \sum_{k=1}^{K}\sum_{t\in A_k}(\rho^*-\bar{R}(b_t,I_t))+\sum_{k=1}^{K}\sum_{t\in E_k}(\rho^*-\bar{R}(b_t,I_t)).
	\end{align}
	
	We next proceed to bound \eqref{reg2}.
	The first term in Equation \eqref{reg2} can be simply upper bounded by:
	\begin{align}\label{reg:explore-bound}
		\sum_{k=1}^{K}\sum_{t\in A_k}(\rho^*-\bar{R}(b_t,I_t)) \leq\sum_{k=1}^{K}\sum_{t\in A_k}\rho^*\leq KI\tau_1\rho^*,
	\end{align}
	where we have used the fact that $\bar{R}$ is non-negative and the length of each exploration phase $A_k$ is $I \tau_1$. Bounding
	the second term in \eqref{reg2} is more delicate, where the length of the exploitation phase $E_k$ is proportional to $\sqrt{k}$. We
	define
	a ``success'' event when the confidence regions contain the true POMDP model, that is, $(\mathcal{P}, {\Omega}) \in \mathcal{C}_k(\delta_k)$ for all $k=1, \ldots, K.$ By the definition of $\delta_k$ in Algorithm \ref{alg:SEEU}, it is easy to see that
	\begin{align}
		\Prob( (\mathcal{P}, \Omega)\notin\mathcal{C}_k(\delta_k), \text{for some}~k)
		\leq \sum_{k=1}^{K}\delta_k=\sum_{k=1}^{K}\frac{\delta}{k^3}\leq \frac{3}{2}\delta.
	\end{align}
	Thus, with probability at least $1-\frac{3}{2}\delta$, the ``success'' event occurs. Then it suffices to bound the regret incurred when the ``success'' event holds in exploitation phases.
	Note when this success event holds,
	we have $\rho^*\leq \rho^k$ for all $k=1, \ldots, K$, where $\rho^*, \rho^k$ are the optimal average reward associated with the true POMDP and the optimistic POMDP in the confidence region $\mathcal{C}_k(\delta_k)$ respectively.
	It follows that, when the ``success'' event holds, we have
	\begin{align}
		&\sum_{k=1}^{K}\sum_{t\in E_k}(\rho^*-\bar{R}(b_t,I_t))\\
		&\leq \sum_{k=1}^{K}\sum_{t\in E_k}(\rho^k-\bar{R}(b_t,I_t))\\
		&=\sum_{k=1}^{K}\sum_{t\in E_k} \left[(\rho^k-\bar{R}(b_t^k,I_t))+(\bar{R}(b_t^k,I_t)-\bar{R}(b_t,I_t)) \right],\label{reg:success-bound-v1}
	\end{align}
	where $b_t=H_{\mathcal{P}, \Omega}^{(t)}(b,i_{0:t-1},o_{1:t})$ and $b_t^k=H_{\mathcal{P}_k, \Omega_k}^{(t)}(b,i_{0:t-1},o_{1:t})$ are the beliefs at time $t$ updated from the initial belief $b$ under the true parameters $(\mathcal{P}, \Omega)$ and the optimistic parameters $(\mathcal{P}_k, \Omega_k)$ from the $k$-th episode under our learning policy.
	
	To bound \eqref{reg:success-bound-v1}, we first bound the second term. Using the definition of $\bar{R}$ in
	\eqref{def:belief-reward-func}, it is straightforward to check that
	\begin{align}
		&\sum_{k=1}^{K} \sum_{t \in E_{k}} \bar{R}\left(b_{t}^{k}, I_{t}\right)-\bar{R}\left(b_{t}, I_{t}\right)\\
		& = \sum_{k=1}^{K} \sum_{t \in E_{k}}\left[\sum_{m \in \mathcal{M}}R(m,I_t) b_{t}^{k}(m)-\sum_{m \in \mathcal{M}}R(m,I_t)b_{t}(m)\right] \\
		&\leq \sum_{k=1}^{K} \sum_{t \in E_{k}} \sum_{m \in \mathcal{M}} R(m,I_t)\left |b_t^k(m)-b_t(m) \right|\\
		& \leq  \sum_{k=1}^{K} \sum_{t \in E_{k}}r_{\max}||b_t^k-b_t||_1.
	\end{align}
	By Proposition \ref{prop:lip_bt}, we have
	\begin{align}
		\Vert b_t -{b}^k_t \Vert_1\leq L_1 \max\limits_{m, i}\left\|\Omega(\cdot|m,i)-\Omega_k(\cdot|m,i)\right\|_1+L_2 \max\limits_{m, i} \left\| P^{(i)}(m,:)-P_k^{(i)}(m,:) \right\|_2,
	\end{align}
	where $L_1=\frac{4(1-\epsilon)^2}{\epsilon^2 \xi}$,
	$L_2=\frac{4(1-\epsilon)^2}{\epsilon^3}$, and $(\mathcal{P}_k , \Omega_k)$ corresponds to the optimistic model in the $k-$th exploitation phase.
	Therefore, the second term in \eqref{reg:success-bound-v1} can be bounded by
	\begin{align}
		&\sum_{k=1}^{K} \sum_{t \in E_{k}} \bar{R}\left(b_{t}^{k}, I_{t}\right)-\bar{R}\left(b_{t}, I_{t}\right) \label{reg:success-bound-p2} \\
		&\leq \sum_{k=1}^{K} \sum_{t \in E_{k}}r_{\max}\left[L_1 \max\limits_{m, i}\left\|\Omega(\cdot|m,i)-\Omega_k(\cdot|m,i)\right\|_1+L_2 \max\limits_{m, i} \left\| P^{(i)}(m,:)-P_k^{(i)}(m,:) \right\|_2\right].\nonumber
	\end{align}

	We leave this here for now and move on to bound the first term in Equation \eqref{reg:success-bound-v1}. We first note that the gain $\rho_k$
	satisfies the Bellman optimality equation for the optimistic belief MDP in $k-$th exploitation phase:
	\begin{align}\label{equ:Bellman-k}
		\rho^k+v_k(b_t^k)= \bar{R}(b_t^k,I_t)+\int_{b_{t+1}^k\in\mathcal{B}}v_k(b_{t+1}^k)\bar{T}_{k}(db_{t+1}^k|b_t^k,I_t)
		=\bar{R}(b_t^k,I_t)+\langle \bar{T}_{k}(\cdot|b_t^k,I_t),v_k(\cdot) \rangle,
	\end{align}
	where $v_k$ is the bias function, $I_t=\pi^{(k)}(b_t^k)$ is the optimal action for $t \in E_k$, and
	$\bar{T}_{k}(\cdot|b_t^k,I_t)=\mathbb{P}_{\mathcal{P}_k,\Omega_k}(b_{t+1}^k\in \cdot|b_t^k,I_t)$ denotes the transition kernel of the belief states conditional on the action under the optimistic POMDP model $(\mathcal{P}_k, \Omega_k)$. Then we can express the first term in \eqref{reg:success-bound-v1} as follows:
	\begin{align}
		&\sum_{k=1}^{K}\sum_{t\in E_k}(\rho^k-\bar{R}(b_t^k,I_t))\\
		&=\sum_{k=1}^{K}\sum_{t\in E_k}(-v_k(b_t^k)+\langle \bar{T}_{k}(\cdot|b_t^k,I_t),v_k(\cdot)\rangle)\\
		&=\sum_{k=1}^{K}\sum_{t\in E_k}(-v_k(b_t^k)+\langle \bar{T}(\cdot|b_t^k,I_t),v_k(\cdot)\rangle)+\langle \bar{T}_{k}(\cdot|b_t^k,I_t)-\bar{T}(\cdot|b_t^k,I_t),v_k(\cdot)\rangle,\label{reg:success-bound-p1}
	\end{align}
	where $\bar{T}(\cdot|b_t^k,I_t)=\mathbb{P}_{\mathcal{P},\Omega}(b_{t+1}^k\in \cdot|b_t^k,I_t)$ denotes the transition kernel of belief states conditional on the action under the true model parameters $(\mathcal{P}, \Omega)$.

	
	
	For the first term in \eqref{reg:success-bound-p1}, we have
	\begin{align}
		&\sum_{k=1}^{K}\sum_{t\in E_k}(-v_k(b_t^k)+\langle \bar{T}(\cdot|b_t^k,I_t),v_k(\cdot)\rangle)\\
		&=\sum_{k=1}^{K}\sum_{t\in E_k}(-v_k(b_t^k)+v_k(b_{t+1}^k))+(-v_k(b_{t+1}^k)+\langle \bar{T}(\cdot|b_t^k,I_t),v_k(\cdot)\rangle)\\
		&=\sum_{k=1}^{K}v_k(b_{t_k+\tau_1I+\tau_2\sqrt{k}}^k)-v_k(b_{t_k+\tau_1I+1}^k)+\sum_{k=1}^{K}\sum_{t\in E_k}\E[v_k(b_{t+1}^k)|\mathcal{F}_{t}]-v_k(b_{t+1}^k).
	\end{align}
	Here, the first term in the last equality is due to the telescoping sum from period $t_k+\tau_1I+1$ to $t_k+\tau_1I+\tau_2\sqrt{k}$, the start and end periods of the exploitation phase in episode $k$. For the second term, if we denote by $\mathcal{F}_{t} \coloneqq \{b_0, I_0, b_1^k, I_1, \cdots, b_{t}^k, I_{t}\}$, then it comes from the relation $\langle \bar{T}(\cdot|b_t^k,I_t),v_k(\cdot)\rangle=\int_{b_{t+1}^k\in\mathcal{B}}v_k(b_{t+1}^k)\bar{T}(db_{t+1}^k|b_t^k,I_t)=\E[v_k(b_{t+1}^k)|b_t^k,I_t]$, by using the Markovian property of the belief sequence.
	
	Applying Proposition \ref{prop:span-uni-bound},	 we have
	\begin{align} \label{eq:bound-span-D}
		v_k(b_{t_k+\tau_1I+\tau_2\sqrt{k}}^k)-v_k(b_{t_k+\tau_1I+1}^k)\leq span(v_k) \le D.
	\end{align}
	We also need the following result, the proof of which relies on a concentration inequality for martingales and  is deferred to the end of this section.
	\begin{lemma}\label{prop:Azuma_belief}
		Let $K$ be the number of total episodes up to time $T$. For each episode $k=1, 2, \cdots, K$, let $E_k$ be the index set of the $k$th exploitation phase, $v_k$ be the relative value function of the optimistic POMDP at the $k$th exploitation phase.
		Then with probability at most $\delta$,
		\begin{align}\label{equ:Azuma}
			\sum_{k=1}^{K}\sum_{t\in E_k}\mathbb{E}[v_k(b_{t+1}^k)|\mathcal{F}_t]-v_k(b_{t+1}^k) \geq D \sqrt{2T\log(\frac{1}{\delta})},
		\end{align}
		where the expectation $\mathbb{E}$ is taken respect to the transition law of belief states $b_t^{k}$ under true parameters $(\mathcal{P},\Omega)$. 
	\end{lemma}
	Applying Lemma \ref{prop:Azuma_belief} and using \eqref{eq:bound-span-D},
	we can infer that with probability at least $1-\delta$, the first term in \eqref{reg:success-bound-p1} satisfies
	\begin{align}\label{reg:success-bound-p1-p1}
		\sum_{k=1}^{K}\sum_{t\in E_k}(-v_k(b_t^k)+\langle \bar{T}(\cdot|b_t^k,I_t),v_k(\cdot)\rangle) \le   KD+D \sqrt{2T\log \left(\frac{1}{\delta}\right)}.
	\end{align}

	We now proceed to bound the second term in Equation \eqref{reg:success-bound-p1}. This part is significantly different from the proof in \cite{jaksch2010near} in that the belief transition $\bar{T}_{k}$ is not directly estimated as the belief itself is not observed. Rather, it depends on the optimistic model $(\mathcal{P}_k, \Omega_k)$ via Equation~\eqref{def:belief-trans-kernel}. We can compute
	\begin{align}
		&\langle \bar{T}_{k}(\cdot|b_t^k,I_t)-\bar{T}(\cdot|b_t^k,I_t),v_k(\cdot)\rangle\\
		&\leq \left|\int_{\mathcal{B}} v_k(b')\bar{T}_k(db'|b_t^k,I_t)-\int_{\mathcal{B}} v_k(b')\bar{T}(db'|b_t^k,I_t)\right|\\
		&=\left| \sum_{o \in \mathcal{O}} v_k\left(H_{k}\left(b_{t}^{k}, I_{t}, o\right) \right)\mathbb{P}_k\left(o| b_{t}^{k}, I_{t}\right)-\sum_{o \in \mathcal{O}} v_k\left(H\left(b_{t}^{k}, I_{t}, o\right) \right)\mathbb{P}\left(o| b_{t}^{k}, I_{t}\right) \right|\\
		& \leq \left| \sum_{o\in \mathcal{O}} v_k\left(H_{k}\left(b_{t}^{k}, I_{t}, o_{t+1}\right) \right) \cdot  \left[\mathbb{P}_k\left(o| b_{t}^{k}, I_{t}\right)- \mathbb{P}\left(o| b_{t}^{k}, I_{t}\right) \right] \right|\\
		&\quad \quad + \left|\sum_{o \in \mathcal{O}} \left[ v_k\left(H_{k}\left(b_{t}^{k}, I_{t}, o\right) \right)-v_k\left(H\left(b_{t}^{k}, I_{t}, o\right) \right)\right]\cdot \mathbb{P}\left(o| b_{t}^{k}, I_{t}\right)\right|,\label{reg:belief-transition-distance}
	\end{align}
	where we use $H_k$ and $H$ to denote the belief updating function under the optimistic model $(\mathcal{P}_k, \Omega_k)$ and the true model $(\mathcal{P}, \Omega)$, and we use $\mathbb{P}_k$ and $\mathbb{P}$ to denote the probability with respect to the optimistic model and true model respectively. To bound the first term in \eqref{reg:belief-transition-distance}, we first note that if the bias function $v_k$ satisfies the Bellman equation \eqref{equ:Bellman-thm}, then so is $v_k+c\bm{1}$ for any constant $c$.
	Thus, without loss of generality, we can assume that $v_k$ satisfies $||v_k||_{\infty}\leq \text{span}(v_k)/2$. Then from Proposition \ref{prop:span-uni-bound}, we have
	$ ||v_k||_{\infty}\leq \frac{1}{2}\text{span}(v_k)\leq \frac{D}{2}.$
	Denote by $O^{(i)}_k(m,o) \coloneqq \Omega_k(o|m,i)$.
	Now we can bound the first term in \eqref{reg:belief-transition-distance} as follows:
	\begin{align}
		&\left| \sum_{o_{t+1} \in \mathcal{O}} v_k\left(H_{k}\left(b_{t}^{k}, I_{t}, o\right) \right) \cdot \left[\mathbb{P}_k\left(o| b_{t}^{k}, I_{t}\right)- \mathbb{P}\left(o| b_{t}^{k}, I_{t}\right) \right] \right| \\
		&\leq \sum_{o\in \mathcal{O}} \left| v_k\left(H_{k}\left(b_{t}^{k}, I_{t}, o\right) \right) \left(\mathbb{P}_k\left(o| b_{t}^{k}, I_{t}\right)- \mathbb{P}\left(o| b_{t}^{k}, I_{t}\right) \right) \right|\\
		&\leq \sum_{o\in \mathcal{O}}\left| \sum_{m'\in\mathcal{M}}\Omega_k(o|m',I_t)\sum_{m\in\mathcal{M}}P^{(I_t)}_k(m,m')b_t^k(m)-\sum_{m'\in\mathcal{M}}\Omega(o|m',I_t)\sum_{m\in\mathcal{M}}P^{(I_t)}(m,m')b_t^k(m)\right| \cdot \| v_k\|_{\infty}\\
		&\leq \left\|(O^{(I_{t})}_k)^{\top}\cdot (P^{(I_t)}_k)^{\top} \cdot b_t^k - (O^{(I_{t})})^{\top} \cdot (P^{(I_t)})^{\top} \cdot b_t^k \right\|_1 \cdot \frac{D}{2}\\
		&\leq \left\|\left(P^{(I_t)}_k \cdot O^{(I_t)}_k \right)^{\top}-\left(P^{(I_t)} \cdot O^{(I_t)} \right)^{\top} \right\|_1 \cdot \left\| b_t^k \right\|_1 \cdot \frac{D}{2}\\
		&= \left\| P^{(I_t)}_k \cdot O^{(I_t)}_k - P^{(I_t)} \cdot O^{(I_t)} \right\|_\infty \cdot \frac{D}{2}\\
		&\leq \left[\left\| P^{(I_t)}_k-P^{(I_t)}\right\|_\infty\cdot \left\| O^{(I_t)}_k \right\|_\infty+ \left\| P^{(I_t)} \right\|_\infty \cdot \left\| O^{(I_t)}_k-O^{(I_t)} \right\|_\infty\right]\cdot \frac{D}{2}\\
		&= \left[\max_m \left\| P^{(I_t)}_k(m,:)-P^{(I_t)}(m,:)\right\|_1+\max_m \left\| O^{(I_t)}_k(m,:)-O^{(I_t)}(m,:)\right\|_1\right] \cdot \frac{D}{2}\\
		&\leq \left[\sqrt{M} \max_{m, i} \left\| P^{(i)}_k(m,:)-P^{(i)}(m,:)\right\|_2+\max_{m,i} \left\|\Omega_k(\cdot|m,i)- \Omega(\cdot|m,i)\right\|_1\right] \cdot \frac{D}{2},\label{reg:belief-transition-distance-p1}
	\end{align}
	where for a matrix, the $L_1$ norm $\|\cdot \|_{1}$ is the maximum of the absolute column sums, and the $L_{\infty}$ norm $\|\cdot \|_{\infty}$ is the maximum of the absolute row sums of the matrix.
	For
	the second term in \eqref{reg:belief-transition-distance}, we can also compute that
	\begin{align}
		&\left|\sum_{o\in \mathcal{O}} \left[ v_k\left(H_{k}\left(b_{t}^{k}, I_{t}, o\right) \right)-v_k\left(H\left(b_{t}^{k}, I_{t}, o\right) \right)\right]\cdot \mathbb{P}\left(o| b_{t}^{k}, I_{t}\right)\right|\\
		& \leq \sum_{o\in \mathcal{O}} \left| v_k\left(H_{k}\left(b_{t}^{k}, I_{t}, o\right) \right)-v_k\left(H\left(b_{t}^{k}, I_{t}, o\right) \right)\right|\cdot \mathbb{P}\left(o| b_{t}^{k}, I_{t}\right)\\
		& \leq \sum_{o \in \mathcal{O}} \frac{D}{2} \left|H_{k}\left(b_{t}^{k}, I_{t}, o\right) -  H\left(b_{t}^{k}, I_{t}, o\right)\right| \cdot \mathbb{P}\left(o| b_{t}^{k}, I_{t}\right)\\
		& \leq \sum_{o \in \mathcal{O}} \frac{D}{2} \left [L_1 \max\limits_{m, i}\left\|\Omega(\cdot|m,I_t)-\Omega_k(\cdot|m,I_t)\right\|_1+L_2 \max\limits_{m, i} \left\| P^{(I_t)}(m,:)-P_k^{(I_t)}(m,:) \right\|_2 \right ] \mathbb{P}\left(o| b_{t}^{k}, I_{t}\right)\\
		&=\frac{D}{2} \left [L_1 \max\limits_{m, i}\left\|\Omega(\cdot|m,I_t)-\Omega_k(\cdot|m,I_t)\right\|_1+L_2 \max\limits_{m, i} \left\| P^{(I_t)}(m,:)-P_k^{(I_t)}(m,:) \right\|_2 \right ],\label{reg:belief-transition-distance-p2}
	\end{align}
	where the second inequality follows from the proof of Proposition \ref{prop:span-uni-bound} (see \eqref{eq:lip-v-k}) that $v_k$ is Lipschitz continuous with Lipschitz module $\frac{D}{2}$, and the third inequality follows from Proposition \ref{prop:lip_bt}.
	On combining \eqref{reg:belief-transition-distance-p1} and \eqref{reg:belief-transition-distance-p2}, we infer that the second term in \eqref{reg:success-bound-p1} can be bounded by
	{\small{  \begin{align}
				&\sum_{k=1}^{K}\sum_{t\in E_k}\langle \bar{T}_{k}(\cdot|b_t^k,I_t)-\bar{T}(\cdot|b_t^k,I_t),v_k(\cdot)\rangle\\
				& \leq \sum_{k=1}^{K}\sum_{t\in E_k}\frac{D}{2} \left [ (1 + L_1) \max\limits_{m, i}\left\|\Omega(\cdot|m,i)-\Omega_k(\cdot|m,i)\right\|_1+ (L_2 +\sqrt{M}) \max\limits_{m, i} \left\| P^{(i)}(m,:)-P_k^{(i)}(m,:) \right\|_2 \right ]. \label{reg:success-bound-p1-p2}
	\end{align} }}
	Together with \eqref{reg:success-bound-p1-p1}, we now obtain a bound for the first term in \eqref{reg:success-bound-v1}:
	\begin{align}
		&\sum_{k=1}^{K}\sum_{t\in E_k}(\rho^k-\bar{R}(b_t^k,I_t))\\
		&\leq  KD+D \sqrt{2T\log \left(\frac{1}{\delta}\right)}
		+\sum_{k=1}^{K}\sum_{t\in E_k} \frac{D}{2} (L_2 + \sqrt{M}) \max\limits_{m, i} \left\| P^{(i)}(m,:)-P_k^{(i)}(m,:) \right\|_2 \\
		&\quad \quad +\sum_{k=1}^{K}\sum_{t\in E_k}\frac{D}{2}(1 +L_1) \max\limits_{m, i}\left\|\Omega(\cdot|m,i)-\Omega_k(\cdot|m,i)\right\|_1.
		\label{reg:success-bound-p1-v2}
	\end{align}
	

	On combining \eqref{reg:success-bound-p1-v2} and \eqref{reg:success-bound-p2}, we can deduce that with probability $1-\delta$, the regret incurred from ``success'' events, i.e., \eqref{reg:success-bound-v1}, can be bounded as follows:
	\begin{align}
		&\sum_{k=1}^{K}\sum_{t\in E_k}(\rho^*-\bar{R}(b_t,I_t))\\
		&\leq KD+D \sqrt{2T\log \left(\frac{1}{\delta}\right)}
		+\sum_{k=1}^{K} \sum_{t \in E_{k}} \left[\frac{D}{2} (L_2 +\sqrt{M}) + r_{\max} L_2 \right] \max\limits_{m, i} \left\| P^{(i)}(m,:)-P_k^{(i)}(m,:) \right\|_2 \\
		&\quad \quad +\sum_{k=1}^{K} \sum_{t \in E_{k}} \left[\frac{D}{2}(1 +L_1) + r_{\max} L_1 \right] \max\limits_{m, i}\left\|\Omega(\cdot|m,i)-\Omega_k(\cdot|m,i)\right\|_1.\label{reg:success-bound-v3}
	\end{align}
	To control the errors between the optimistic model $(\mathcal{P}_k, \Omega_k)$ and the true model $(\mathcal{P}, \Omega)$ in the above equation, we use the confidence intervals in Proposition \ref{prop:spectral} where the sample size needs to be appropriately large. To this end,
	denote $T_0 \le T$ the time period that the number of samples collected in the exploration phases for action $i$ to exceed $N_0^{(i)}$ given in Proposition \ref{prop:spectral}, for all $i \in \mathcal{I}$. Also let $k_0$ be the episode that $T_0$ is in.
	Under the confidence level $\delta_k=\delta/k^3$, we can infer from Proposition \ref{prop:spectral} that on the ``success" event, for $k \ge k_0$
	\begin{eqnarray}
		\left\|\Omega(\cdot|m,I_t)-\Omega_k(\cdot|m,I_t)\right\|_1 &\leq C_1 \sqrt{\frac{\log\left(\frac{6k^3(O^2+O)}{\delta}\right)}{\tau_1 k}}, \label{eq:CI1}\\
		\left\|P^{(I_t)}(m,:)- P_k^{(I_t)}(m,:)\right\|_2&\leq C_2\sqrt{\frac{\log\left(\frac{6k^3(O^2+O)}{\delta}\right)}{\tau_1 k}}. \label{eq:CI2}
	\end{eqnarray}
	Thus, formula \eqref{reg:success-bound-v3}, or the second term of \eqref{reg2}, can be bounded as follows:
	{  \small{    \begin{align}
				&\sum_{k=1}^{K}\sum_{t\in E_k}(\rho^*-\bar{R}(b_t,I_t))\\
				&\leq KD+D \sqrt{2T\log \left(\frac{1}{\delta}\right)}+T_0\rho^*\\
				&\quad +\sum_{k=k_0}^{K}\tau_2\sqrt{k}\left[\left(\frac{D}{2}+\frac{D}{2}L_1 + r_{\max} L_1\right)C_1+\left(\frac{D\sqrt{M}}{2}+ +\frac{D}{2}L_2+r_{\max} L_2 \right)C_2 \right]\sqrt{\frac{\log(\frac{6(O^2+O)k^3}{\delta})}{\tau_1 k}}\\
				&\leq KD+D \sqrt{2T\log \left(\frac{1}{\delta}\right)}+T_0\rho^*\\
				&\quad+K\tau_2\left[\left(\frac{D}{2}+\frac{D}{2}L_1 + r_{\max} L_1\right)C_1+\left(\frac{D\sqrt{M}}{2}+\frac{D}{2}L_2+r_{\max} L_2 \right)C_2 \right]\sqrt{\frac{\log(\frac{6(O^2+O)K^3}{\delta})}{\tau_1}}.\label{reg:success-bound-v4}
	\end{align} } }
	Note the ``success'' event holds with probability $1-\frac{3}{2}\delta$, and the bound in Lemma \ref{prop:Azuma_belief} holds with probability $\delta$. Hence together with \eqref{reg:explore-bound}, we can get that with probability at least $1-\frac{5}{2}\delta$, the following bound holds for the regret in \eqref{reg2}:
	{  \small{     \begin{align} \label{reg:filtration}
				&\sum_{t=1}^{T}\rho^*-\bar{R}(b_t,I_t)\\
				&\leq K(\tau_1I\rho^*+D)+D \sqrt{2T\log \left (\frac{1}{\delta}\right)}+T_0\rho^*\\
				&\quad\quad+K\tau_2\left[\left(\frac{D}{2}+\frac{D}{2}L_1 + r_{\max} L_1\right)C_1+\left(\frac{D\sqrt{M}}{2}+\frac{D}{2}L_2+r_{\max} L_2 \right)C_2 \right]\sqrt{\frac{\log(\frac{6(O^2+O)K^3}{\delta})}{\tau_1}}. \nonumber
	\end{align} }}
	On combining \eqref{reg:expreward-reward} and \eqref{reg:filtration}, we can infer that with probability at least $1-\frac{7}{2}\delta$, the regret in \eqref{reg} can be bounded by
	\begin{align}
		&(T+1)\rho^* - \sum_{t=0}^T R_t^{\pi}(b)\\
		&\leq K(\tau_1I\rho^*+D)+D \sqrt{2T \log\left(\frac{1}{\delta}\right)}+\sqrt{2Tr_{\max}\log \left(\frac{1}{\delta}\right)}+T_0\rho^*\\
		&\quad\quad+K\tau_2\left[\left(\frac{D}{2}+\frac{D}{2}L_1 + r_{\max} L_1\right)C_1+\left(\frac{D\sqrt{M}}{2} +\frac{D}{2}L_2+r_{\max} L_2 \right)C_2 \right]\sqrt{\frac{\log(\frac{6(O^2+O)K^3}{\delta})}{\tau_1}}.
	\end{align}
	It remains to bound the number of episodes $K.$
	Note that
	\begin{align}
		\sum_{k=1}^{K-1}(\tau_1I+\tau_2\sqrt{k}) \leq T \leq \sum_{k=1}^{K}(\tau_1I+\tau_2\sqrt{k}),
	\end{align}
	so the number of episodes $K$ is bounded by $(\frac{T}{\tau_1I+\tau_2})^{2/3}\leq K\leq 3(\frac{T}{\tau_2})^{2/3}$ .
	It follows that with probability at least
	$1-\frac{7}{2}\delta$,
	\begin{align}
		(T+1)\rho^* - \sum_{t=0}^T R_t^{\pi}(b)& \leq CT^{2/3}\sqrt{\log\left(\frac{9(O+1)}{\delta}T\right)}+T_0\rho^*,
	\end{align}
	where
	\begin{align}
		C=&3\sqrt{2}\left[\left(\frac{D}{2}+\frac{D}{2}L_1 + r_{\max} L_1\right)C_1+\left(\frac{D\sqrt{M}}{2}+ \frac{D}{2}L_2+r_{\max} L_2 \right)C_2 \right]\tau_1^{-1/2}\tau_2^{1/3}\\
		&\quad\quad+3(\tau_1I\rho^*+D)\tau_2^{-2/3}+D \sqrt{2 \log\left(\frac{1}{\delta}\right)}+\sqrt{2r_{\max}\log \left(\frac{1}{\delta}\right)}.\label{para:C}
	\end{align}
	Thus, the regret defined in \eqref{def:reg} can be bounded
	\begin{align}
		\mathcal{R}_T^{\pi}=\max_{b}\left\{(T+1)\rho^* - \sum_{t=0}^T R_t^{\pi}(b)\right\} \leq CT^{2/3}\sqrt{\log\left(\frac{9(O+1)}{\delta}T\right)}+T_0\rho^*.
	\end{align}
	The proof is therefore complete. \Halmos
	
	
	\subsection{Proof of Lemma~\ref{prop:Azuma_belief}}
	Let $E=\cup_{k=1}^K E_k$ be the set of all exploitation time steps up to time $T$, and $\bar{N}=\sum_{i=0}^T \1_{i\in E}$ be the total number of periods in the exploitation phases, and $v_k$ be the value function of the optimistic POMDP at the $k$th exploitation phase.
	Define a stochastic process $\{Z_{n}: n=0,1, \ldots, \bar{N}\}$:
	\begin{align}
		&Z_0=0,\\
		&Z_{n}=\sum_{j=1}^{n} \mathbb{E}[v_{k_j}(b_{t_j+1}^{k_j})|\mathcal{F}_{t_j}]-v_{k_j}(b_{t_j+1}^{k_j}), \quad n \geq 1,
	\end{align}
	where $t_j:=\min\{t:\sum_{i=0}^t\1_{i \in E}=j\}$ denotes the time step corresponding to the $j-$th period in the exploitation phases,
	and
	$k_j:=\{k:t_j \in E_k\}$. 
	
	
	We first show that $\{Z_{n}\}$ is a martingale.
	Note that $\E[|Z_{n}|]\leq \sum_{j=1}^{n} \text{span}(v_{k_j})\leq nD\le TD<\infty$ by Proposition~\ref{prop:span-uni-bound}. Let $\bar{\mathcal{F}}_n \coloneqq \mathcal{F}_{t_n} = \{b_0, I_0, b_1^k, I_1, \cdots, b_{t_n}^k, I_{t_n}\}$. Then we can check that
	\begin{align}
		\E[Z_{n}-Z_{n-1}|\bar{\mathcal{F}}_{n-1}]=\E[\mathbb{E}[v_{k_n}(b_{t_n+1}^{k_n})|\mathcal{F}_{t_n}]-v_{k_n}(b_{t_n+1}^{k_n})|\bar{\mathcal{F}}_{n-1}]=0,
	\end{align}
	where the last equality is due to $\bar{\mathcal{F}}_{n-1}=\mathcal{F}_{t_{n-1}} \subset \mathcal{F}_{t_n}$ and then applying the tower property of conditional expectations. Hence
	$\{Z_n \}$ is a martingale, adapted to the filtration $\{\mathcal{F}_n\}$.
	Moreover, by Proposition~\ref{prop:span-uni-bound}, we have
	\begin{align}
		|Z_{n}-Z_{n-1}|=|\mathbb{E}[v_{k_n}(b_{t_n+1}^{k_n})|\mathcal{F}_{t_n}]-v_{k_n}(b_{t_n}^{k_n})| \leq \text{span}(v_{k_n}) \leq D.
	\end{align}
	Thus, $\{Z_n\}$ is a martingale with bounded difference.
	
	Applying the Azuma-Hoeffding inequality \citep{azuma1967weighted}, we have
	\begin{align}
		\mathbb{P}(Z_{\bar{N}}-Z_{0} \geq \epsilon) \leq \exp \left(\frac{-\epsilon^{2}}{2 \bar{N} D^{2}}\right).
	\end{align}
	Note that $\bar{N} \leq T$ and $Z_{\bar{N}}=\sum_{k=1}^{K}\sum_{t\in E_k}\mathbb{E}[v_k(b_{t+1}^k)|\mathcal{F}_{t}]-v_k(b_{t+1}^k)$. Thus, setting $\epsilon=D \sqrt{2T\log\left(\frac{1}{\delta}\right)}$, we can obtain
	\begin{align}
		\mathbb{P}\left(\sum_{k=1}^{K}\sum_{t\in E_k}\mathbb{E}[v_k(b_{t+1}^k)|\mathcal{F}_{t}]-v_k(b_{t+1}^k) \geq D \sqrt{2T\log\left(\frac{1}{\delta}\right)}\right) \leq \delta.
	\end{align}
	The proof is complete.\Halmos

	\section{Regret analysis of the ETC algorithm}\label{sec:etc}
	\rev{ In this section, we discuss the ETC (Explore-then-Commit) algorithm for learning POMDPs and its regret analysis. The detailed steps of the algorithm are summarized below in Algorithm~\ref{alg:etc}. The structure of the ETC algorithm is the same as the SEEU algorithm. The main difference is that in each exploitation phase, the ETC algorithm uses the point estimator of the POMDP parameters directly while the SEEU algorithm uses the optimistic estimator in the confidence region.}
	\rev{
		\begin{algorithm}[h!]
			\caption{Explore-then-Commit Algorithm}
			\label{alg:etc}
			\begin{algorithmic}[1]
				\REQUIRE Exploration hyperparameter $\tau_1$,  exploitation hyperparameter $\tau_2$.
				\STATE Initialize: time $T_1=0$, initial belief $b_0$.
				\FOR {$k=1,2,3,\dots$}
				\FOR{$t=T_k,T_k+1,\dots,T_k+\tau_1I$}
				\STATE Select each action $\tau_1$ times successively.
				\STATE Observe next observation $o_{t+1}$.
				\ENDFOR
				\STATE Use the realized actions and observations in all previous exploration phases $\hat{\mathcal{I}}_k \coloneqq \{i_{T_1:T_1+\tau_1I}, \cdots,i_{T_k:T_k+\tau_1I}
				\}$ and $\hat{\mathcal{O}}_k \coloneqq \{o_{T_1+1:T_1+\tau_1I+1},\cdots,o_{T_k+1:T_k+\tau_1I+1}
				\}$ as input to Algorithm~\ref{alg:spectral} to compute\\
				\centerline{$(\hat{\mathcal{P}}_k, \hat{\Omega}_k) = \textbf{SpectralEstimation}(\hat{\mathcal{I}}_k, \hat{\mathcal{O}}_k)$.}
				\FOR{$t=0,1,\dots,T_k+\tau_1I$}
				\STATE Update belief $b_t^k$ to $ b_{t+1}^k=H_{\hat{\mathcal{P}}_k, \hat{\Omega}_k}(b_t^k,i_t,o_{t+1})$ under parameters $(\hat{\mathcal{P}}_k, \hat{\Omega}_k)$.
				\ENDFOR
				\FOR{$t=T_k+\tau_1I+1,\dots,T_k+\tau_1I+\tau_2\sqrt{k}$}
				\STATE Execute the optimal policy $\pi^{(k)}$ by solving the Bellman equation \eqref{equ:Bellman-thm} with parameters $(\hat{\mathcal{P}}_k, \hat{\Omega}_k)$: $i_t=\pi_t^{(k)}(b_t^k)$.
				\STATE Observe next observation $o_{t+1}$.
				\STATE Update the belief at $t+1$ following
				$b_{t+1}^k=H_{\hat{\mathcal{P}}_k, \hat{\Omega}_k}(b_t^k,i_t,o_{t+1})$.
				\ENDFOR
				\STATE $T_{k+1}\gets t+1$
				\ENDFOR
			\end{algorithmic}
		\end{algorithm}
	}
	
	\subsection{Sensitivity analysis of average-reward POMDPs}
	\rev{   To establish regret bounds for the ETC algorithm, we first analyze the sensitivity of the optimal average-reward of the (undiscounted) POMDP with respect to the model parameters.
	}
	
	\rev{
		To this end, we first consider two distinct infinite-horizon discounted POMDPs with identical rewards and identical discount factor $\beta \in (0,1)$, with different model parameters $\theta = (\mathcal{P},\Omega)$  and $\hat \theta = (\mathcal{ \hat P}, \hat \Omega)$. Let $\mu^*(\theta)$ and $\mu^*(\hat \theta)$ denote the optimal policies of these two POMDPs. Let $J_{\mu^*(\theta)}(b, \theta)$ and $J_{\mu^*(\hat \theta)}(b, \hat \theta)$ respectively denote the optimal discounted rewards of these two POMDPs with initial belief state $b$. Then from Theorem 14.9.1 in \cite{krishnamurthy2016partially}, we obtain
		\begin{align} \label{eq:inf-discounted-sensitivity}
			\left| J_{\mu^*(\theta)}(b, \theta) - J_{\mu^*(\hat \theta)}(b, \hat \theta) \right| \le 3 K ||\theta - \hat \theta||,
		\end{align}
		where $K = \frac{r_{\max}}{1 - \beta}$ and $||\theta - \hat \theta|| = \max_{m, i } \sum_{m', o } \left| P^{(i)}(m, m') \Omega(o|m', i) -  \hat P^{(i)}(m, m') \hat  \Omega(o|m', i) \right|$. It is easy to see that
		\begin{align}
			||\theta - \hat \theta|| &= \max_{m, i } \sum_{m', o } \left| P^{(i)}(m, m') \Omega(o|m', i) -  \hat P^{(i)}(m, m') \hat  \Omega(o|m', i) \right| \\
			& \le  \max_{m, i } \sum_{m', o } \left| P^{(i)}(m, m') \Omega(o|m', i) -  \hat P^{(i)}(m, m')  \Omega(o|m', i) \right| \\
			& \quad +  \max_{m, i } \sum_{m', o } \left| \hat P^{(i)}(m, m') \Omega(o|m', i) -  \hat P^{(i)}(m, m') \hat  \Omega(o|m', i) \right| \\
			& \le  \max_{m, i } \sum_{m' } \left| P^{(i)}(m, m')  -  \hat P^{(i)}(m, m') \right| +  \max_{m, i } \sum_{m', o } \left| \Omega(o|m', i) -  \hat  \Omega(o|m', i) \right| \cdot \left| \hat P^{(i)}(m, m') \right| \\
			&=  \max_{m, i } || P^{(i)}(m, :)  -  \hat P^{(i)}(m, :) ||_1 +  \max_{m, i } \sum_{m' } || \Omega(\cdot |m', i) -  \hat  \Omega(\cdot |m', i) ||_1 \cdot \hat P^{(i)}(m, m')  \\
			&\le  \max_{m, i } || P^{(i)}(m, :)  -  \hat P^{(i)}(m, :) ||_1 +  \max_{m', i } || \Omega(\cdot |m', i) -  \hat  \Omega(\cdot |m', i) ||_1.
		\end{align}
		Equation~\eqref{eq:inf-discounted-sensitivity} gives the sensitivity of the optimal infinite-horizon discounted reward with respect to model parameters. To study the the sensitivity of average-reward POMDP, we use the vanishing discount factor method. From the proof of Proposition~\ref{lemma:exist-optimal-policy} (see Equation~\ref{eq:bound-v-beta}), we know that when $\theta = (\mathcal{P},\Omega)$ satisfy Assumptions~\ref{assum:trans_matrix_min} and \ref{assum:reward_density_min}, the bias functions for the infinite-horizon discounted problems are bounded uniformly in the discount factor. With this key condition verified, we can infer that $\lim_{\beta \rightarrow 1-} (1-\beta) J_{\mu^*(\theta)}(b, \theta) = \rho^*(\theta)$ for all $b$. See also Equation~\eqref{eq:v-limit}. When for instance the purturbation $||\theta - \hat \theta||$ is small, $\hat \theta = (\mathcal{ \hat P}, \hat \Omega) $ will also satisfy Assumptions~\ref{assum:trans_matrix_min} and \ref{assum:reward_density_min} with $\epsilon$ and $\xi$ replaced by $\epsilon/2$ and $\xi/2$ respectively.
		Then, we can multiply $1-\beta$ at both sides of Equation~\eqref{eq:inf-discounted-sensitivity}, send $\beta$ to one, and obtain
		\begin{align} \label{eq: inf-undiscounted-sensitivity}
			\left| \rho^*(\theta) - \rho^*(\hat \theta) \right| \le 3 r_{\max} \cdot ||\theta - \hat \theta||,
		\end{align}
		where $\rho^*(\theta)$ and $\rho^*(\hat \theta)$ respectively denote the optimal average reward of the two POMDPs.
	}
	\subsection{Regret analysis of the ETC algorithm}
	
	\rev{  In this section, we analyze the regret of the ETC algorithm, where point estimators are used in the exploitation phases instead of optimistic estimators. The regret analysis of the ETC algorithm is similar to the analysis of the SEEU algorithm given in the proof of Theorem~\ref{thm:upper_bound}, however, we need the sensitivity of average-reward POMDP in \eqref{eq: inf-undiscounted-sensitivity}. In the analysis below, we only highlight the necessary changes.
		Specifically, for optimistic estimators we have used the following inequality \eqref{reg:success-bound-v1} on the success event when the confidence region contains the true model:
		\begin{align}
			\sum_{k=1}^{K}\sum_{t\in E_k}(\rho^*-\bar{R}(b_t,I_t))
			\leq \sum_{k=1}^{K}\sum_{t\in E_k}( \rho^k-\bar{R}(b_t,I_t)),
		\end{align}
		where $\rho^*$ and $\rho^k$ are the optimal average reward associated with the true POMDP and the optimistic POMDP in the confidence region $\mathcal{C}_k(\delta_k)$ respectively, and $\rho^*\leq \rho^k$ by optimism. For point estimators, 
		we can directly use the inequality
		\begin{align}
			\sum_{k=1}^{K}\sum_{t\in E_k}(\rho^*-\bar{R}(b_t,I_t))
			\leq \sum_{k=1}^{K}\sum_{t\in E_k}(\hat \rho^k-\bar{R}(b_t,I_t)) + \sum_{k=1}^{K}\sum_{t\in E_k}( \rho^* - \hat \rho^k) ,
		\end{align}
		where $\hat \rho^k$ denotes the optimal average reward of the estimated POMDP at episode $k$. The same arguments to bound $\sum_{k=1}^{K}\sum_{t\in E_k}( \rho^k-\bar{R}(b_t,I_t))$ can be used to bound the term $\sum_{k=1}^{K}\sum_{t\in E_k}(\hat \rho^k-\bar{R}(b_t,I_t)),$ because optimism is not used in these arguments. Hence by using the point estimator in the exploitation phase, we only need to bound the extra term $ \sum_{k=1}^{K}\sum_{t\in E_k}( \rho^* - \hat \rho^k)$.  As in the proof of Theorem~\ref{thm:upper_bound}, denote by $T_0$ the time period that the number of samples collected in the exploration phases for action $i$ to exceed $N_0^{(i)}$ given in Proposition \ref{prop:spectral}, for all $i \in \mathcal{I}$. Let $k_0$ be the episode that $T_0$ is in. It suffices to bound the  term $ \sum_{k=k_0}^{K}\sum_{t\in E_k}( \rho^* - \hat \rho^k)$ because the regret incurred before episode $k_0$ is simply bounded by the constant $T_0 \rho^*$. 
		From the proof of Theorem~\ref{thm:upper_bound} we know that with probability $1- \frac{3}{2} \delta$, for all $k_0 \le k \le K,$
		\begin{eqnarray}
			\left\|\Omega(\cdot|m, i )- \hat \Omega_k(\cdot|m,i)\right\|_1 &\leq C_1 \sqrt{\frac{\log\left(\frac{6(O^2+O) K^3}{\delta}\right)}{\tau_1 k}}, \\
			\left\|P^{(i)}(m,:)- \hat P_k^{(i)}(m,:)\right\|_2&\leq C_2\sqrt{\frac{\log\left(\frac{6 (O^2+O) K^3}{\delta}\right)}{\tau_1 k}},
		\end{eqnarray}
		for all $i \in \mathcal{I}$ and $m \in \mathcal{M},$ where $(\hat P_k,  \hat  \Omega_k)$ denotes the point estimators in episode $k.$
		By the sensitivity analysis of the optimal average reward in \eqref{eq: inf-undiscounted-sensitivity}, we have
		\begin{align}
			& \sum_{k=k_0}^{K}\sum_{t\in E_k}( \rho^* - \hat \rho^k) \\
			& \le \sum_{k=k_0}^{K}\sum_{t\in E_k} 3 r_{\max} \left( \max_{m, i } || P^{(i)}(m, :)  -  \hat P_k^{(i)}(m, :) ||_1 +  \max_{m', i } || \Omega(\cdot |m', i) -  \hat  \Omega_k(\cdot |m', i) ||_1 \right)\\
			& \le   \sum_{k=k_0}^{K}\sum_{t\in E_k} 3 r_{\max} \left( \max_{m, i } \sqrt{M} || P^{(i)}(m, :)  -  \hat P_k^{(i)}(m, :) ||_2 +  \max_{m', i } || \Omega(\cdot |m', i) -  \hat  \Omega_k(\cdot |m', i) ||_1 \right).
		\end{align}
		Since the length of the exploitation phase $E_k$ is $\tau_2 \sqrt{k}$, we can deduce that with probability $1- \frac{3}{2} \delta,$
		\begin{align}
			\sum_{k=k_0}^{K}\sum_{t\in E_k}( \rho^* - \hat \rho^k) \le
			3 r_{\max} K\tau_2\left[ C_1+C_2 \sqrt{M}\right]\sqrt{\frac{\log(\frac{6(O^2+O)K^3}{\delta})}{\tau_1}}.
		\end{align}
		Since $(\frac{T}{\tau_1I+\tau_2})^{2/3}\leq K\leq 3(\frac{T}{\tau_2})^{2/3}$, we can follow the proof of Theorem~\ref{thm:upper_bound} to obtain the following result which shows that the regret of the ETC algorithm is also of $O(T^{2/3} \sqrt{\log T})$.
		\begin{theorem}\label{thm:upper_bound_ETC}
			Fix the hyperparameter $\tau_1$ in the ETC algorithm to be sufficiently large. Suppose Assumptions 1 to 4 hold.
			There exist a constant $T_0$ such that for $T>T_0$, with probability at least $1-\frac{7}{2}\delta$ ,
			the regret of the ETC Algorithm satisfies
			\begin{align}
				\mathcal{R}_T& \leq CT^{2/3}\sqrt{\log\left(\frac{9(O+1)}{\delta}T\right)}+ T_0\rho^* + 9 r_{\max} T^{2/3}  \left( C_1+C_2 \sqrt{M}\right) \sqrt{\log\left(\frac{9(O+1)T}{\delta}\right)} \tau_1^{-1/2}\tau_2^{1/3},
			\end{align}
			where $\rho^* \le r_{\max}$ and the constant $C$ is given in \eqref{const: C}.
		\end{theorem}
	}
	

\end{document}